\documentclass[preprint,1p]{elsarticle}

\newif\ifarxiv
\arxivtrue

\ifarxiv
\else
\fi

\usepackage{lineno} %
\usepackage{arydshln}
\usepackage[hyphens]{url}
\usepackage[colorlinks=true,allcolors=black]{hyperref}
\modulolinenumbers[5]

\usepackage{xspace} %
\usepackage{textcomp} %
\usepackage{gensymb} %
\usepackage{arydshln}
\usepackage{booktabs}
\usepackage{multirow}
\usepackage[utf8]{inputenc}
\usepackage[T1]{fontenc}
\usepackage[acronym]{glossaries}
\usepackage[caption=false,farskip=0pt]{subfig}
\usepackage[table,dvipsnames]{xcolor}

\usepackage[misc]{ifsym} %
\usepackage{diagbox}
\usepackage[capitalise,noabbrev]{cleveref} %

\newcommand{\plus}{\texttt{+}}

\usepackage{pifont} %
\newcommand{\cmark}{\ding{51}}%
\newcommand{\xmark}{\ding{55}}%

\newcounter{fncounter}
\newcommand\customfootnote[1]{\stepcounter{fncounter}\footnote{\hspace{0.3mm}#1}}

\newcounter{fnpytorch}
\newcounter{fnkeras}
\newcounter{fndarknet}

\newcounter{fnImagesInternet}

\newcommand*{\RL}[2][]{\textcolor{Rhodamine}{[\textbf{\ifthenelse{\equal{#1}{}}{RL}{RL(#1)}}: #2]}}
\newcommand*{\DM}[2][]{\textcolor{orange}{[\textbf{\ifthenelse{\equal{#1}{}}{DM}{DM(#1)}}: #2]}}
\newcommand*{\VE}[2][]{\textcolor{green}{[\textbf{\ifthenelse{\equal{#1}{}}{VE}{VE(#1)}}: #2]}}

\newcommand*{\GM}[2][]{\textcolor{Turquoise}{[\textbf{\ifthenelse{\equal{#1}{}}{GM}{GM(#1)}}: #2]}}

\newcommand*{\todo}[2][]{\textcolor{red}{[\textbf{\ifthenelse{\equal{#1}{}}{TODO}{TODO(#1)}}: #2]}}

\newcommand\red[1]{{\textcolor{red}{#1}}}

\newcommand\major[1]{#1} %

\newacronymstyle{long-short-br}
{%
  \GlsUseAcrEntryDispStyle{long-short}%
}%
{%
  \GlsUseAcrStyleDefs{long-short}%
}
\setacronymstyle{long-short-br}

\journal{IET Intelligent Transport Systems}

\biboptions{authoryear}

\ifarxiv
    \makeatletter
    \def\ps@pprintTitle{%
      \let\@oddhead\@empty
      \let\@evenhead\@empty
      \let\@oddfoot\@empty
      \let\@evenfoot\@oddfoot
    }
    \makeatother

    \usepackage{transparent}
    \usepackage{tikz}
    \newcommand\copyrighttext{%
      \scriptsize This is the author-prepared version of an article published in \textit{IET Intelligent Transport Systems} (DOI: \href{https://doi.org/10.1049/itr2.70086}{\textcolor{blue}{10.1049/itr2.70086}}).}
    \newcommand\copyrightnotice{%
    \begin{tikzpicture}[remember picture,overlay]
    \node[anchor=south,yshift=95pt,xshift=0pt] at (current page.south) {\fbox{\transparent{0.85}\parbox{\dimexpr0.575\textwidth-\fboxsep-\fboxrule\relax}{\copyrighttext}}};
    \end{tikzpicture}%
    }
\else
\fi

\begin{document}

\newacronym{alpr}{ALPR}{Automatic License Plate Recognition}
\newacronym{amr}{AMR}{Automatic Meter Reading}
\newacronym{cnn}{CNN}{Convolutional Neural Network}
\newacronym{crnn}{CRNN}{Convolutional Recurrent Neural Network}
\newacronym{ctc}{CTC}{Connectionist Temporal Classification}
\newacronym{denatran}{DENATRAN}{National Traffic Department of Brazil}
\newacronym{fps}{FPS}{Frames Per Second}
\newacronym{gan}{GAN}{Generative Adversarial Network}
\newacronym{grcnn}{GRCNN}{Gated Recurrent Convolution Neural Network}
\newacronym{lp}{LP}{license plate}
\newacronym{lpd}{LPD}{License Plate Detection}
\newacronym{lpr}{LPR}{License Plate Recognition}
\newacronym{lodo}{LODO}{leave-one-dataset-out}
\newacronym{lstm}{LSTM}{Long Short-Term Memory}
\newacronym{iou}{IoU}{Intersection over Union}
\newacronym{it}{IT}{information technology}
\newacronym{nme}{NME}{Normalization Mean Error}
\newacronym{ocr}{OCR}{Optical Character Recognition}
\newacronym{rare}{RARE}{Robust text recognizer with Automatic REctification}
\newacronym{rodosol}{RodoSol}{\textit{Rodovia do Sol}}
\newacronym{rtwoam}{R\textsuperscript{2}AM}{Recursive Recurrent neural networks with Attention Modeling}
\newacronym{sgd}{SGD}{Stochastic Gradient Descent}
\newacronym{starnet}{STAR-Net}{SpaTial Attention Residue Network}
\newacronym{trba}{TRBA}{TPS-ResNet-BiLSTM-Attention}

\newcommand{\aolp}{AOLP\xspace}
\newcommand{\caltech}{Caltech Cars\xspace}
\newcommand{\ccpd}{CCPD\xspace}
\newcommand{\cdhard}{CD-HARD\xspace}
\newcommand{\chineselp}{ChineseLP\xspace}
\newcommand{\clpd}{CLPD\xspace}
\newcommand{\englishlp}{EnglishLP\xspace}
\newcommand{\openalprbr}{OpenALPR-BR\xspace}
\newcommand{\openalpreu}{OpenALPR-EU\xspace}
\newcommand{\pku}{PKU\xspace}
\newcommand{\rodosol}{RodoSol-ALPR\xspace}
\newcommand{\rodosolalpr}{\rodosol}
\newcommand{\ssigsegplate}{SSIG-SegPlate\xspace}
\newcommand{\stills}{UCSD-Stills\xspace}
\newcommand{\ufop}{UFOP\xspace}
\newcommand{\ufpralpr}{UFPR-ALPR\xspace}

\newcommand{\dataset}{GAN-ALPR\xspace}

\newcommand{\crnn}{\acrshort*{crnn}\xspace}
\newcommand{\grcnn}{\acrshort*{grcnn}\xspace}
\newcommand{\rare}{\acrshort*{rare}\xspace}
\newcommand{\rosetta}{Rosetta\xspace}
\newcommand{\rtwoam}{\acrshort*{rtwoam}\xspace}
\newcommand{\starnet}{\acrshort*{starnet}\xspace}
\newcommand{\trba}{\acrshort*{trba}\xspace}
\newcommand{\vitstrbase}{ViTSTR-Base\xspace}
\newcommand{\vitstrsmall}{ViTSTR-Small\xspace}
\newcommand{\vitstrtiny}{ViTSTR-Tiny\xspace}
\newcommand{\vitstr}{ViTSTR\xspace}

\newcommand{\holistic}{Holistic-CNN\xspace}
\newcommand{\multitask}{Multi-Task\xspace}
\newcommand{\multitaskLR}{Multi-Task-LR\xspace}
\newcommand{\cnng}{CNNG\xspace}

\newcommand{\crnet}{CR-NET\xspace}
\newcommand{\fastocr}{Fast-OCR\xspace}

\newcommand{\yolocsp}{YOLOv4-CSP\xspace}
\newcommand{\scaledyolo}{\yolocsp}
\newcommand{\lpnme}{LP-NME\xspace}
\newcommand{\cyclegan}{CycleGAN\xspace}
\newcommand{\pixtwopix}{pix2pix\xspace}
\newcommand{\stargan}{StarGAN\xspace}
\newcommand{\xception}{Xception\xspace}
\newcommand{\cdcc}{CDCC-NET\xspace}
\newcommand{\cdccnet}{\cdcc}
\newcommand{\iwpod}{IWPOD-NET\xspace}
\newcommand{\iwpodnet}{\iwpod}
\newcommand{\smallerlocatenet}{Smaller-LocateNet\xspace}
\newcommand{\smallerlocate}{\smallerlocatenet}
\newcommand{\locatenet}{LocateNet\xspace}
\newcommand{\hybridmobilenet}{Hybrid-MobileNetV2\xspace}

\newcommand{\numDatasets}{12\xspace}
\newcommand{\numModels}{16\xspace}
\newcommand{\numTimesTrained}{9\xspace}
\newcommand{\numTimesTrainedWords}{nine\xspace}

\newcommand{\supplementarySplits}{\url{https://raysonlaroca.github.io/supp/lpr-synthetic-data/splits.zip}}

\newcommand{\minRR}{94.6\xspace}
\newcommand{\maxRR}{97.9\xspace}

\newcommand{\avgRR}{94.9\xspace}
\newcommand{\avgRectRR}{96.4\xspace}

\newcommand{\bestRRnoDataAug}{93.7\xspace}
\newcommand{\bestRR}{\maxRR}

\newcommand{\avgOrigPermutation}{91.4\xspace}
\newcommand{\avgOrigPermutationRect}{93.6\xspace}
\newcommand{\avgOrigTemplates}{92.5\xspace}
\newcommand{\avgOrigTemplatesRect}{94.7\xspace}
\newcommand{\avgOrigGAN}{93.2\xspace}
\newcommand{\avgOrigGANRect}{95.2\xspace}

\newcommand{\OursCCPDCLPD}{97.3\xspace}
\newcommand{\OursCCPDPKU}{99.5\xspace}

\ifarxiv
\else
    \thispagestyle{empty}
    \pagestyle{empty}
    
    \noindent \textbf{Title of the manuscript:} Advancing Multinational License Plate Recognition Through Synthetic and Real Data Fusion: A Comprehensive Evaluation
    
    \vspace{2mm}
    
    \noindent \textbf{All the authors' complete names:} Rayson Laroca, Valter Estevam, Gladston Juliano Prates Moreira, Rodrigo Minetto, David Menotti
    
    \vspace{2mm}
    
    \noindent \textbf{Departmental \& institutional affiliations of the authors:}
    \begin{itemize}
        \item Affiliation 1: Graduate Program in Informatics, Pontifical Catholic University of Paraná, Curitiba, Brazil
        \item Affiliation 2: Graduate Program in Informatics, Federal University of Paraná, Curitiba, Brazil
        \item Affiliation 3: Federal Institute of Paraná, Irati, Brazil
        \item Affiliation 4: Computing Department, Federal University of Ouro Preto, Ouro Preto, Brazil
        \item Affiliation 5: Department of Computing, Federal University of Technology-Paraná, Curitiba, Brazil
    \end{itemize}
    
    \noindent \textbf{Postal \& e-mail address of the corresponding author}:
    \begin{itemize}
        \item Bloco 8 – Parque Tecnológico – 2º andar, Rua Imaculada Conceição, 1155 - Prado Velho, 80215-901, Curitiba, Brasil
        \item rayson@ppgia.pucpr.br
    \end{itemize}
    
    \vspace{2mm}
    
    \noindent \textbf{Conflict of interest statement:} The authors declare that they have no conflict of interest.
    
    \vspace{2mm}
    
    \noindent \textbf{Funding information:} This work was supported by the Coordination for the Improvement of Higher Education Personnel~(CAPES) and the National Council for Scientific and Technological Development~(CNPq) (grant number~315409/2023-1).
    
    \vspace{2mm}
    
    \noindent \textbf{Data availability statement:} The data that support the findings of this study are openly available at \url{https://raysonlaroca.github.io/supp/lpr-synthetic-data/}.
    
    \vspace{2mm}
    
    \noindent \textbf{Credit contribution statement:}
    \begin{itemize}
      \item \textbf{Rayson Laroca}: Conceptualization, Data curation, Formal analysis, Investigation, Methodology, Software, Validation, Visualization, Writing – original draft
      \item \textbf{Valter Estevam}: Conceptualization, Investigation, Writing – original draft, Writing – review \& editing
      \item \textbf{Gladston J. P. Moreira}: Conceptualization, Investigation, Writing – review \& editing
      \item \textbf{Rodrigo Minetto}: Conceptualization, Investigation, Writing – review \& editing, Supervision
      \item \textbf{David Menotti}: Conceptualization, Investigation, Writing – review \& editing, Supervision, Resources, Project administration, Funding acquisition
    \end{itemize}
    
    \clearpage
\fi

\begin{frontmatter}

\title{Advancing Multinational License Plate Recognition Through Synthetic and Real Data Fusion: A Comprehensive Evaluation}

\ifarxiv
    \author[pucpr,ufpr]{Rayson~Laroca\corref{corr}}
    \cortext[corr]{Corresponding author}
    \ead{rayson@ppgia.pucpr.br}
    \author[ufpr,ifpr]{Valter Estevam}
    \author[ufop]{Gladston J. P. Moreira}
    \author[utfpr]{\\Rodrigo Minetto}
    \author[ufpr]{David~Menotti}
\else
    \author[pucpr,ufpr]{Rayson~Laroca\corref{corr}}
    \cortext[corr]{Corresponding author}
    \ead{rayson@ppgia.pucpr.br}
    \author[ufpr,ifpr]{Valter Estevam}
    \ead{valter.junior@ifpr.edu.br}
    \author[ufop]{Gladston J. P. Moreira}
    \ead{gladston@ufop.edu.br}
    \author[utfpr]{\\Rodrigo Minetto}
    \ead{rminetto@utfpr.edu.br}
    \author[ufpr]{David~Menotti}
    \ead{menotti@inf.ufpr.br}
\fi

\address[pucpr]{Pontifical Catholic University of Paraná (PUCPR), Curitiba, Brazil}
\address[ufpr]{Federal University of Paran\'a (UFPR), Curitiba, Brazil}
\address[ifpr]{Federal Institute of Paran\'a (IFPR), Irati, Brazil}
\address[ufop]{Federal University of Ouro Preto (UFOP), Ouro Preto, Brazil}
\address[utfpr]{Federal University of Technology-Paran\'{a} (UTFPR), Curitiba, Brazil}

\begin{abstract}
\acrlong*{alpr} \major{is} a frequent research topic due to its wide-ranging practical applications.
While recent studies \major{use} synthetic images to improve \gls*{lpr} results, there remain several limitations in these efforts.
This work addresses these constraints by comprehensively exploring the integration of real and synthetic data to enhance \gls*{lpr} performance.
We subject \numModels~\gls*{ocr} models to a benchmarking process involving \numDatasets~public datasets acquired from various regions.
Several key findings \major{emerge} from our investigation.
Primarily, the massive incorporation of synthetic data substantially \major{boosts} model performance in both intra- and cross-dataset scenarios.
We \major{examine} three distinct methodologies for generating synthetic data: template-based generation, character permutation, and utilizing a \gls*{gan} model, each contributing significantly to performance enhancement.
The combined use of these methodologies \major{demonstrates} a notable synergistic effect, leading to end-to-end results that \major{surpass} those reached by state-of-the-art methods and established commercial systems.
Our experiments also \major{underscore} the efficacy of synthetic data in mitigating challenges posed by limited training data, enabling remarkable results to be achieved even with small fractions of the original training data.
Finally, we \major{investigate} the trade-off between accuracy and speed among different models, identifying those that strike the optimal balance in each intra-dataset and cross-dataset~settings.

\end{abstract}


\end{frontmatter}

\ifarxiv
    \copyrightnotice
\else
\fi

\pagenumbering{arabic}
\pagestyle{plain}

\section{Introduction}
\label{sec:introduction}

\glsresetall

\gls*{alpr} systems employ image processing and pattern recognition techniques to locate and recognize \glspl*{lp} in images or videos~\citep{laroca2021efficient,fan2022improving,rao2024license}.
Traffic law enforcement, toll collection, and vehicle access control in restricted areas are some practical applications for an \gls*{alpr} system~\citep{weihong2020research,ding2024endtoend}.

In the deep learning era, \gls*{alpr} systems typically include two stages: \gls*{lpd} and \gls*{lpr}~\citep{li2018reading,xu2018towards,ke2023ultra}.
The former refers to locating the \gls*{lp} regions in the input image, while the latter refers to identifying the characters on each \gls*{lp}.
Recent research has focused on the \gls*{lpr} stage~\citep{liu2021fast,nascimento2023super,schirrmacher2023benchmarking}, as generic object detectors (e.g., YOLO) have achieved impressive results in the \gls*{lpd} stage for some time now~\citep{hsu2017robust,xie2018new,hu2020mobilenet}.

Increased mobility and internationalization set new challenges for developing effective \gls*{alpr} systems, as they must handle \glspl*{lp} from multiple regions with non-standardized formats~\citep{lubna2021automatic,trinh2022layout}.
While \gls*{alpr} systems have exhibited remarkable performance on \glspl*{lp} from diverse regions (e.g., Brazil, mainland China, Europe, Taiwan, among others) due to advances in deep learning and the increasing availability of annotated datasets~\citep{henry2020multinational,silva2022flexible,liu2024irregular}, recent studies have indicated the existence of strong biases in \gls*{alpr} research.
An example worth mentioning is~\citep{laroca2022first}, where the authors showed that each dataset has a unique and identifiable ``signature,'' as a lightweight classification model could predict the source dataset of an \gls*{lp} image at levels significantly better than~chance.

One way to mitigate the problem of biases in \gls*{alpr} would be to embrace the ``wildness'' of the internet to collect a large-scale dataset from multiple sources~\citep{torralba2011unbiased,laroca2022first}.
However, labeling such a dataset would be very expensive and time-consuming~\citep{bjorklund2019robust,han2020license,gao2023group}, not to mention the growing concerns surrounding privacy~\citep{chan2020european,kong2021federated,trinh2023pp4av}.
In this scenario, synthetic data emerges as a practical alternative, offering a cost-effective, privacy-preserving solution while providing the diversity and scale needed to train deep learning-based models~effectively.

\major{Although recent research has explored creating synthetic \gls*{lp} images to improve \gls*{lpr} performance, our analysis in \cref{sec:related_work} reveals certain limitations in these efforts.
Existing studies have predominantly employed a single methodology to generate synthetic \glspl*{lp}, leaving unanswered questions regarding the potential for significantly enhanced outcomes by integrating data generated from various methodologies.
Additionally, most works have focused on \glspl*{lp} from a single region, even though this limitation has been acknowledged for many years in the literature~\citep{mecocci2006generative,anagnostopoulos2008license}.
To illustrate, researchers have trained separated instances of generative models --~e.g., \glspl*{gan}~-- for different \gls*{lp} layouts.
This approach becomes increasingly impractical and even unfeasible as the number of \gls*{lp} layouts the \gls*{alpr} system must handle increases.
Ultimately, the assessment of synthetic data generation methods has primarily relied on the performance of individual \gls*{ocr} models, overlooking the fact that images created using a particular method may disproportionately favor certain models over~others.}

This work aims to address the limitations described above by delving further into the integration of real and synthetic data to enhance \gls*{lpr}.
Setting our research apart from previous studies, we subject $\numModels$ well-known \gls*{ocr} models to a benchmarking process across \numDatasets public datasets acquired from multiple regions.
Synthetic \gls*{lp} images are created by drawing inspiration from the three most widely adopted methodologies in the literature: a rendering-based pipeline (templates), character permutation, and a \gls*{gan} model.
We conduct ablation studies to demonstrate the impact of each methodology on the final results and the importance of synthetic data when training data is~scarce.

In summary, this paper makes the following contributions:
\vspace{-1.5mm}
\begin{itemize}
    \item The most extensive experimental evaluation conducted in the field.
    While our focus lies on the \gls*{lpr} stage, as per recent research trends, we also compare various models for detecting the \glspl*{lp} and their corresponding corners within the input images.
    Our end-to-end experiments cover both intra- and cross-dataset evaluations, including an examination of the speed/accuracy trade-off of the \gls*{ocr}~models;
    \vspace{-1.5mm}
    \item \major{We deviate from prior methodologies by introducing a pipeline that employs a single \gls*{gan} model to generate images of \glspl*{lp} from diverse regions and across styles.
    Notably, satisfactory outcomes are attained despite using a relatively small number of real images for training.
    This success stems from our approach of supplementing these real images with many synthetic ones created by a different method while also leveraging an \gls*{ocr} model to identify and filter out poorly generated~images;}
    \vspace{-1.5mm}
    \item Our results show that the massive use of synthetic data significantly improves the performance of the models, both in intra- and cross-dataset scenarios.
    Remarkably, employing the top-performing \gls*{ocr} model yields end-to-end results surpassing state-of-the-art methods and established commercial systems.
    These findings are particularly impressive because our models \major{are} not specifically trained for any particular \gls*{lp} layout, and we do not rely on post-processing with heuristic rules to improve the \gls*{lpr} performance on \glspl*{lp} from specific~regions;
    \vspace{-1.5mm}
    \item Our ablation studies reveal that each synthesis method contributes considerably to enhancing the results, with a substantial synergistic effect observed when combining them.
    Incorporating synthetic data into the training set also proves to be effective in overcoming the challenges posed by limited training data, as commendable results are attained even when using only small fractions of the original~data;
    \vspace{-1.5mm}
    \item We will publicly release all synthetic images created for training the \gls*{ocr} models, along with the accompanying code, thus enabling the generation of new~images.
\end{itemize}

\major{While recent studies have introduced new recognition architectures to boost LPR accuracy, our work adopts a complementary, data-centric approach. We show that significant performance improvements can be achieved across a wide range of architectures simply by enhancing the quality of the training data. This strategy supports fair, architecture-agnostic comparisons and emphasizes reproducibility and generalizability by providing plug-and-play training data that can be leveraged by any \gls*{ocr}~model.}

The remainder of this paper is structured as follows.
\cref{sec:related_work} outlines the prevalent methods for synthesizing \gls*{lp} images in the literature.
\cref{sec:synthetic} elaborates on our methodology for generating synthetic data, which \major{are} integrated with real data to train the \gls*{ocr} models.
\cref{sec:experimental-setup} describes the experimental setup, including the datasets and models explored.
The results are presented and analyzed in \cref{sec:results}.
Finally, \cref{sec:conclusions} summarizes the key findings of this~study.

\section{Related Work}
\label{sec:related_work}

In addition to the dataset bias mentioned in the previous section, \gls*{lpr} faces challenges related to unbalanced data.
The inherent difficulty in collecting \gls*{lp} images from a variety of regions makes most \gls*{alpr} datasets exhibit a significant bias towards specific regional identifiers~\citep{zhang2021robust_attentional,liu2021fast,wang2022efficient,shvai2023multiple}.

Considering the above discussion, many methods have been proposed to generate synthetic \gls*{lp} images.
These methods aim to mitigate bias in the experiments and minimize the reliance on large volumes of real images for training \gls*{ocr} models.
The subsequent paragraphs provide a concise overview of three widely adopted~methods.

A highly intuitive approach for creating \gls*{lp} images involves a rendering-based process, particularly effective as \glspl*{lp} within a specific region typically conform to a strict standard.
Put simply, such a method initiates with a blank template mirroring the actual aspect ratio and color scheme of \glspl*{lp} from the target region.
Then, a random sequence of characters reflecting the actual \gls*{lp} sequence scheme is superimposed onto the template.
Finally, transformations are applied to enhance the diversity of the generated~images.

Several works have effectively explored the above methodology, including but not limited to~\citep{bjorklund2019robust,maier2022reliability,gao2023group}.
Regarding the creation of \gls*{lp} images, these works primarily differed in the \gls*{lp} layout synthesized and the specific transformations applied.
For instance, \cite{bjorklund2019robust} focused on creating Italian \glspl*{lp}, \cite{maier2022reliability} generated German \glspl*{lp}, and \cite{gao2023group} synthesized \glspl*{lp} from mainland China.
In general, the transformations applied include modifications in font thickness, pixel shifts in character positions, \gls*{lp} rotation, adjustments in brightness and contrast, and the introduction of random shadows and~noise.

Rendering-based methods face a significant limitation as they generate images with inconsistent distributions compared to real-world images, even when incorporating many transformations~\citep{wu2019pixtextgan,maier2022reliability,gao2023group}.
Consequently, \gls*{lpr} models trained solely on such images often produce unsatisfactory outcomes when applied to real-world images.
Taking this into account, researchers have explored various approaches for creating realistic \gls*{lp} images, ranging from simpler methods such as character permutation to more complex strategies involving generative models (see~below).

Generating synthetic data through character permutation is a simple yet effective method for achieving balance among character classes.
Considering that each character's position on a given \gls*{lp} is labeled, one character can be replaced by another by superimposing the corresponding patch.
Typically, this procedure focuses on replacing overrepresented characters in the training set with those underrepresented.
To our knowledge, \cite{goncalves2018realtime} were the first to explore this permutation-based approach in the \gls*{lpr} context.
Since then, several authors have successfully applied it to construct well-balanced training sets regarding character classes.
The following paragraph presents three examples and briefly describes the subtle variations in how the respective authors implemented this~method.

\cite{shashirangana2022license} swapped character patches from distinct \gls*{lp} images, while most authors limited their permutations to character patches from the same \gls*{lp} to reduce illumination inconsistencies. 
\cite{albatat2022end} refrained from permuting patches of thin characters such as `1' and `I' to prevent potential deformation caused by swapping them with wider characters.
In contrast, other authors addressed this issue by first expanding the bounding boxes of smaller characters, incorporating portions of the \gls*{lp} background into them, to ensure uniform sizing of all characters before permutation.
Lastly, although most authors swapped letters with digits and vice versa, \cite{laroca2023leveraging} only permuted letters with letters and digits with digits, enabling models to implicitly learn the fixed positions for letters and digits in certain \gls*{lp}~layouts.

Concerning the use of generative models in \gls*{lpr} research, the prevailing choice has been \glspl*{gan}.
The application of conditional \glspl*{gan} to image-to-image translation was first investigated by~\cite{isola2017image}, with the proposal of the widely recognized pix2pix model.
Pix2pix learns to map an image from the input to the output domain using an adversarial loss in conjunction with the L1 loss between the output and target images, thus requiring paired training data.
While paired image-to-image translation models have shown remarkable results since this seminal work, acquiring such training data (i.e., matching image pairs with pixel-wise or patchwise labeling) can be time-consuming or even unrealistic.
To tackle this challenge, \cyclegan, DualGAN and DiscoGAN provided a novel perspective (nearly simultaneously), in which the models discover relations between two visual domains without any explicitly paired data.
As paired data is often unavailable, unpaired image-to-image translation has gained much attention in subsequent years. %
In the following paragraphs, we briefly describe recent publications that have employed \glspl*{gan} to generate synthetic data for improved~\gls*{lpr}.

\cite{wang2022efficient} employed \cyclegan~\citep{zhu2017unpaired} to transform a large number of script \gls*{lp} images, created using OpenCV, into realistic ones.
Implementation details were not provided.
Similarly, \citet{zhang2021robust_attentional} trained \cyclegan without the second cycle-consistency loss (i.e., they discarded the loss responsible for mapping real images into synthetic ones) to generate \gls*{lp} images with different characters and distinct characteristics.
They trained multiple networks, each specialized in producing images with specific attributes.
For instance, one model was trained to transform script images into bright \glspl*{lp}, while another was trained to convert script images into dark \glspl*{lp}, and so forth.
In both works, \glspl*{lp} of only a few different styles (all from mainland China) were synthesized.
\cite{fan2022improving} adopted essentially the same approach but trained \cyclegan with the Wasserstein distance loss.
Their experiments focused solely on two distinct \gls*{lp} styles, one from mainland China and the other from the Taiwan~region.

\citet{han2020license} trained \cyclegan, \stargan and pix2pix to generate images of the prevalent style of Korean \glspl*{lp} from script images.
Their findings indicated that \pixtwopix produced more realistic and diverse \gls*{lp} images, supported by both qualitative comparisons and the superior performance of an \gls*{ocr} model trained with {\pixtwopix}-generated images compared to instances of the same model trained with images from \cyclegan and \stargan.
\cite{shashirangana2022license} employed pix2pix to convert color images from the \ccpd dataset into infrared images.
They explored the KAIST multi-spectral dataset, which has $95$k paired color and infrared images, for training the pix2pix model.
The researchers suggested that the generated images could be employed to train an \gls*{ocr} model capable of identifying \glspl*{lp} extracted from real images captured during nighttime periods.
\cite{shvai2023multiple} built on several existing frameworks (e.g., AC-GAN and PG-GAN) to generate high-quality \gls*{lp} images with distinct sequences.
In summary, their model achieves diversity by inputting the generator with different random latent vectors.
It is worth noting that the authors focused on generating a single style of \glspl*{lp}, specifically the most common style found on vehicles in Texas, United~States.

When examining the works described in this section, as well as others omitted for brevity, it becomes clear that the evaluation of methods for generating synthetic data has relied on the outcomes produced by individual \gls*{ocr} models.
For example, \cite{wang2022efficient} assessed the efficacy of their strategy solely based on the results achieved by their \gls*{cnn}-based model.
Similarly, \cite{zhang2021robust_attentional} considered only the results reached by an \gls*{ocr} model based on \xception, and \cite{fan2022improving} considered only the results yielded by \cnng, their multi-task model.
We posit that such an evaluation is suboptimal because images created through a specific method may disproportionately benefit certain approaches over others.
This phenomenon was evidenced in \citep{laroca2019convolutional}, where two segmentation-free models (\multitask and \crnn) had a much higher performance gain than the YOLO-based \crnet model~\citep{silva2020realtime} when incorporating images generated via character permutation into the training set.
Therefore, there is a lack of studies focused on evaluating these techniques' efficiency based on the results achieved by multiple \gls*{ocr} models with varying~characteristics.

Another point that catches our attention is that most works still focus on \glspl*{lp} from a single region.
In fact, it is not uncommon for only a very specific \gls*{lp} style (e.g., single-row blue \glspl*{lp} from mainland China) to be considered in the experiments~\citep{han2020license,maier2022reliability,shvai2023multiple}.
Researchers have often opted to train separate instances of the proposed models for each \gls*{lp} layout.
For example, one model generates/recognizes \glspl*{lp} from the Taiwan region, another model generates/recognizes \glspl*{lp} from mainland China, and so forth~\citep{bjorklund2019robust,zhang2021efficient,wang2022rethinking}.
As mentioned earlier, this approach becomes increasingly impractical and even unfeasible as the number of \gls*{lp} layouts the \gls*{alpr} system must handle increases.
This impracticality arises from the necessity of adjusting parameters and retraining models whenever incorporating support for \glspl*{lp} from new regions or even different \gls*{lp} styles within the same~region.

Ultimately, it is crucial to emphasize that within the examined literature, each work has exclusively generated synthetic \glspl*{lp} through a single methodology, such as relying solely on templates, employing only character permutation, or using \glspl*{gan} exclusively.
It remains unclear whether relying on a single approach is sufficient for optimal results, or if considerably superior outcomes could be attained by integrating data generated through diverse methodologies.
This study addresses this gap by benchmarking $\numModels$ \gls*{ocr} models across \numDatasets publicly available datasets from different regions.
We explore all three methodologies mentioned earlier for generating synthetic data, which are elaborated on in the following section.
To our knowledge, this represents the most extensive experimental evaluation conducted in the \gls*{alpr}~field.

\section{Synthetic Data}
\label{sec:synthetic}

This section details our approach to generate synthetic data, which \major{is} combined with real data to train the deep models for \gls*{lpr}.
We start by outlining the methodology for creating \gls*{lp} images using blank templates and character patches sourced from the internet.
Afterward, we delve into the process of producing new \gls*{lp} images by permuting the positions of the characters within each \gls*{lp}.
Lastly, we elaborate on our utilization of a paired image-to-image translation model (pix2pix) to generate realistic \gls*{lp}~images.

\subsection{Templates}
\label{sec:synthetic:templates}

\setcounter{fnImagesInternet}{\thefncounter}

\begin{figure}[!htb]
    \centering
    
    \resizebox{0.9\linewidth}{!}{
    \includegraphics[height=5.5ex]{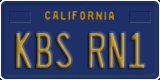}
    \includegraphics[height=5.5ex]{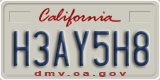}
    \includegraphics[height=5.5ex]{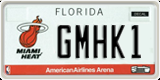}
    \includegraphics[height=5.5ex]{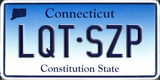}
    \includegraphics[height=5.5ex]{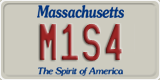}
    \includegraphics[height=5.5ex]{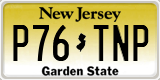}
    \includegraphics[height=5.5ex]{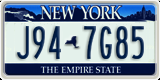}
    }
    
    \vspace{0.6mm}
    
    \resizebox{0.9\linewidth}{!}{
    \includegraphics[height=5.5ex]{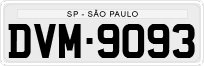}
    \includegraphics[height=5.5ex]{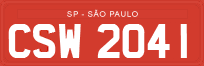}
    \includegraphics[height=5.5ex]{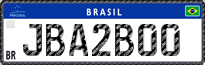}
    \includegraphics[height=5.5ex]{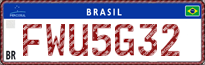}
    \includegraphics[height=5.5ex]{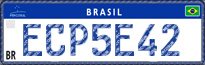}
    }
    
    \vspace{0.6mm}
    
    \resizebox{0.9\linewidth}{!}{
    \includegraphics[height=5.25ex]{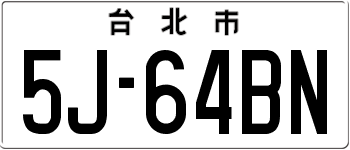}
    \includegraphics[height=5.25ex]{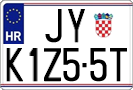}
    \includegraphics[height=5.25ex]{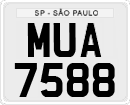}
    \includegraphics[height=5.25ex]{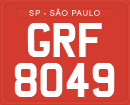}
    \includegraphics[height=5.25ex]{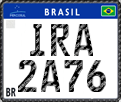}
    \includegraphics[height=5.25ex]{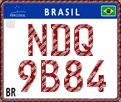}
    \includegraphics[height=5.25ex]{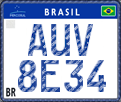}
    \includegraphics[height=5.25ex]{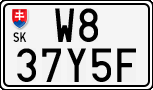}
    \includegraphics[height=5.25ex]{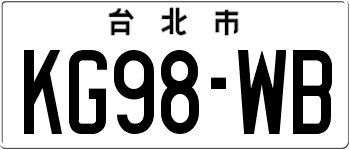}
    }
    
    \vspace{0.6mm}
    
    \resizebox{0.9\linewidth}{!}{
    \includegraphics[height=4.5ex]{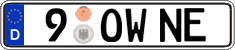}
    \includegraphics[height=4.5ex]{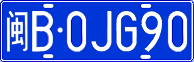}
    \includegraphics[height=4.5ex]{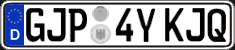}
    \includegraphics[height=4.5ex]{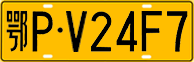}
    \includegraphics[height=4.5ex]{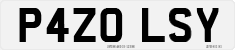}
    }
    
    \vspace{0.6mm}
    
    \resizebox{0.9\linewidth}{!}{
    \includegraphics[height=4.5ex]{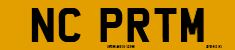}
    \includegraphics[height=4.5ex]{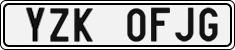}
    \includegraphics[height=4.5ex]{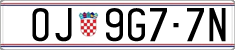}
    \includegraphics[height=4.5ex]{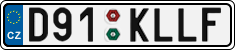}
    }

    \vspace{-2mm}
    
    \caption{%
    Examples of the template-based \gls*{lp} images we created for this work.
    Notably, any sequence can be generated for each template.
    The background and character images were gathered from the internet$^\thefnImagesInternet$.
    During training, these LP images are subjected to various transformations to introduce~variability.
    }
    \label{fig:samples-templates}
\end{figure}

While there are various approaches for creating \gls*{lp} images using templates, the method employed in this study is quite straightforward.
First, blank templates that match the aspect ratio and color scheme of real \glspl*{lp} are sourced from the internet\customfootnote{Most of the blank templates and character patches were taken from \url{https://platesmania.com/}}.
Subsequently, a sequence of characters, selected randomly yet crafted to mirror the patterns found on authentic \glspl*{lp}, is superimposed onto each template.
\cref{fig:samples-templates} shows examples of \gls*{lp} images generated through this process.
Naturally, during the training of the \gls*{ocr} models, we subject these images to various transformations to introduce variability.
These transformations encompass a range of techniques, including but not limited to random perspective transformation, introduction of random noise, incorporation of random shadows, and application of random changes to hue, saturation and~brightness.

To better simulate real-world scenarios, the templates we \major{generate} using this method \major{are} derived from the \gls*{lp} styles observed within the training sets of the datasets explored in our experiments (see \cref{sec:experiments:datasets}).
In other words, we \major{do not create} templates for \gls*{lp} styles found exclusively in the test sets.
To illustrate, one of the datasets we \major{employ} in our cross-dataset assessments contains images of electric vehicles registered in mainland China, which feature 8-character green \glspl*{lp}.
Despite this, we \major{refrain} from creating templates for this \gls*{lp} style since it is not present in the training~set.

An appealing aspect of this synthesis method lies in its ability to generate any sequence for each template while adhering to a predefined number of characters.
Nevertheless, two drawbacks deserve attention.
First, as highlighted in \cref{sec:related_work}, images produced by such rendering-based approaches often exhibit inconsistent distributions compared to real-world images.
Second, sourcing background and character images online for certain \gls*{lp} styles, particularly those less popular or recently introduced, can pose a challenge.
This challenge \major{plays} a key role in our decision not to create templates for every \gls*{lp} style present in the training set, in addition to the inherent scope limitations of our~study.

We \major{generate} $100$k \gls*{lp} images employing this approach, a number determined through preliminary experiments that showed slightly improved outcomes compared to using $50$k images and similar performance to using $200$k images.
The number of synthesized \glspl*{lp} \major{is} balanced across the six explored \gls*{lp} layouts (i.e., American, Brazilian, Chinese, European, Mercosur, and Taiwanese)\customfootnote{As in previous works, the ``Chinese'' layout refers to \glspl*{lp} assigned to vehicles registered in mainland China, while the ``Taiwanese'' layout denotes \glspl*{lp} issued for vehicles registered in the Taiwan~region.}, and the \gls*{lp} sequences \major{are} defined to maximize class balance for each character~position.

\subsection{Character Permutation}
\label{sec:synthetic:permutation}

Generating synthetic data through character permutation is also a straightforward process, outlined as follows.
Initially, each character's bounding box ($x$, $y$, $w$, $h$) must be labeled.
Then, if all the bounding boxes share the same width and height, the patch of each character can be replaced with another according to predefined rules.
However, it is important to highlight that characters from distinct classes often differ in size, especially in terms of width.
Adhering to established practices in the literature (refer to \cref{sec:related_work}), we first \major{expand} the bounding boxes of smaller characters, incorporating small portions of the \gls*{lp} background into them, so that all characters have identical dimensions.
Subsequently, we \major{replace} patches of characters that \major{are} overrepresented in the training set with patches from those that were underrepresented.
To maintain consistency in illumination, we \major{limit} character permutation to patches within the same~\gls*{lp}.

\begin{figure}[!htb]
    \centering
    
    \resizebox{0.55\linewidth}{!}{ %
    \includegraphics[height=4.5ex]{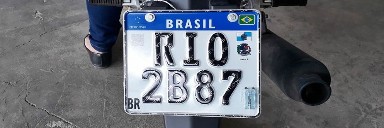} \hspace{-1.9mm}
    \includegraphics[height=4.5ex]{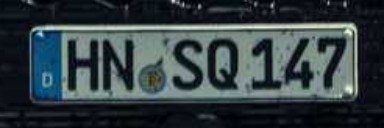} \hspace{-1.9mm}
    \includegraphics[height=4.5ex]{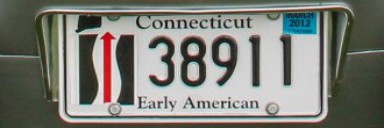}
    } %
    
    \vspace{0.1mm}
    
    \resizebox{0.55\linewidth}{!}{ %
    \includegraphics[height=4.5ex]{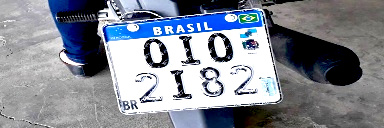} \hspace{-1.9mm}
    \includegraphics[height=4.5ex]{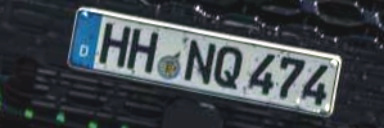} \hspace{-1.9mm}
    \includegraphics[height=4.5ex]{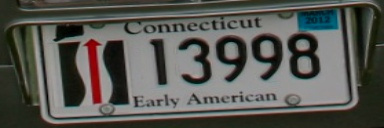}
    } %
    
    \vspace{0.1mm}
    
    \resizebox{0.55\linewidth}{!}{ %
    \includegraphics[height=4.5ex]{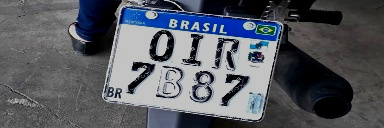}  \hspace{-1.9mm}
    \includegraphics[height=4.5ex]{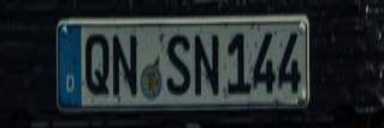} \hspace{-1.9mm}
    \includegraphics[height=4.5ex]{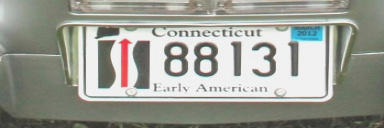}
    } %
    
    \vspace{0.1mm}
    
    \resizebox{0.55\linewidth}{!}{ %
    \includegraphics[height=4.5ex]{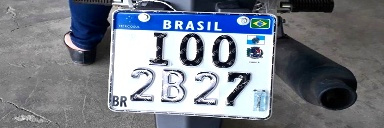} \hspace{-1.9mm}
    \includegraphics[height=4.5ex]{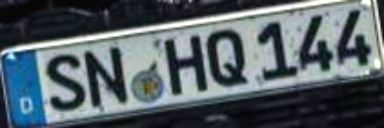} \hspace{-1.9mm}
    \includegraphics[height=4.5ex]{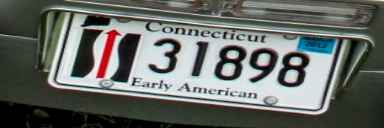} 
    } %
    
    \vspace{-2mm}
    
    \caption{
    Some \gls*{lp} images created by permuting the positions of the characters within each \gls*{lp} and then applying transformations.
    The images in the top row are the originals, while the others were~synthesized.}
    \label{fig:samples-permutation}
\end{figure}

In \cref{fig:samples-permutation}, we show examples of \gls*{lp} images generated by permuting the character positions on three \glspl*{lp} and applying random transformations of scale, rotation, brightness and cropping.
Despite the impressive visual outcomes, it is essential to acknowledge certain limitations associated with this image synthesis method.
First, manually labeling the bounding box for each character on every \gls*{lp} image is a laborious, time-consuming, and error-prone task~\citep{bjorklund2019robust,wang2022rethinking,liu2024irregular}.
Second, this method can only be applied to \gls*{lp} images where the character bounding boxes do not intersect (typically restricting its use on tilted \glspl*{lp}).
Otherwise, parts of some characters may become obscured or replicated during the permutation process.
Lastly, as the permutations involve repetitions and are limited to characters within the same~\gls*{lp}, the \gls*{ocr} models may inadvertently learn undesirable correlations or biases.
For instance, \cite{goncalves2018realtime} pointed out that characters from initially underrepresented classes exhibited a strong self-correlation, as they are more likely to appear in multiple positions on the permuted \glspl*{lp} (this is illustrated in \cref{fig:samples-permutation} as~well).

We conducted a series of experiments in the validation set to determine the number of \gls*{lp} images to generate through this approach.
We then generated $300$k images, evenly distributed across the different \gls*{lp} layouts, as we found that generating a higher volume of images did not yield improved~results.

\subsection{Image-To-Image Translation (pix2pix)}
\label{sec:synthetic:dataset}

\major{As outlined in \cref{sec:related_work}, most previous works explored unpaired image-to-image translation methods (e.g., \cyclegan) to generate realistic \gls*{lp} images due to the lack of labeled paired data.
In this work, we exploit the character permutation method described above to tackle this problem.
More specifically, we generated over one million new \gls*{lp} images by shuffling the character positions on approximately $2$k images from the training set of public datasets and the internet.}
While \cite{laroca2021efficient} provided labels for most of these images, we further enriched the annotations by labeling the positions of the \gls*{lp} corners.
The complete set of annotations will be publicly~available.

\major{Considering that these images are accompanied by precise annotations for the position of each \gls*{lp} corner and the bounding box of every character, they can be used to train paired image-to-image translation methods.}
In this study, we employ the renowned pix2pix model~\citep{isola2017image} for synthesizing many realistic images of \glspl*{lp} from multiple regions.
We remark that although there are newer models available that would certainly yield better results than pix2pix, our decision to opt for pix2pix is primarily based on its widespread availability across various frameworks such as Chainer, Keras, PyTorch, TensorFlow, Torch, and others\customfootnote{See a list of pix2pix implementations at \url{https://phillipi.github.io/pix2pix/}. Our chosen implementation can be found at~\url{https://github.com/affinelayer/pix2pix-tensorflow}.}.
This choice \major{is} particularly significant for our research, given that part of our experiments were conducted on an old CPU lacking AVX instructions, significantly limiting the available framework~options.

\begin{figure}[!htb]
	\captionsetup[subfigure]{labelformat=empty,captionskip=0.5pt}
	\centering
			        
	\resizebox{0.85\linewidth}{!}{
		\subfloat[input]{
			\includegraphics[height=6.5ex]{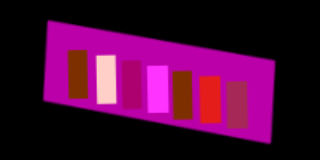}
			} \hspace{-3mm}
		\subfloat[output]{
			\includegraphics[height=6.5ex]{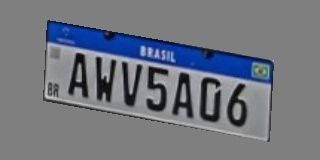}
			} \hspace{0.1mm}
		\subfloat[input]{
			\includegraphics[height=6.5ex]{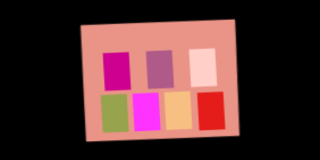}
			} \hspace{-3mm}
		\subfloat[output]{
			\includegraphics[height=6.5ex]{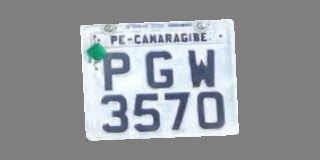}
			} \hspace{0.1mm}
		\subfloat[input]{
			\includegraphics[height=6.5ex]{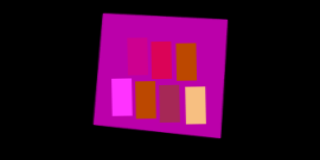}
			} \hspace{-3mm}
		\subfloat[output]{
			\includegraphics[height=6.5ex]{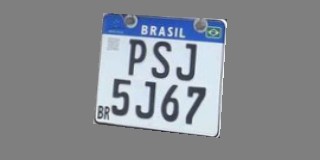}
		}
	}
			        
	\vspace{1.5mm}
			        
	\resizebox{0.85\linewidth}{!}{
		\subfloat[input]{
			\includegraphics[height=4.5ex]{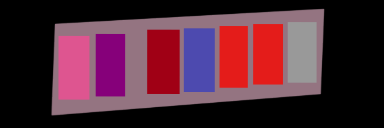}
			} \hspace{-3mm}
		\subfloat[output]{
			\includegraphics[height=4.5ex]{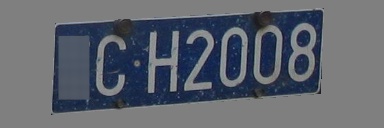}
			} \hspace{0.1mm}
		\subfloat[input]{
			\includegraphics[height=4.5ex]{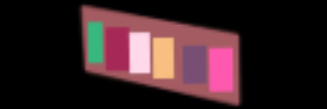}
			} \hspace{-3mm}
		\subfloat[output]{
			\includegraphics[height=4.5ex]{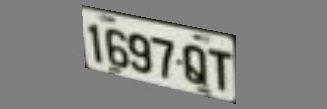}
			} \hspace{0.1mm}
		\subfloat[input]{
			\includegraphics[height=4.5ex]{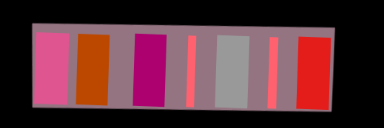}
			} \hspace{-3mm}
		\subfloat[output]{
			\includegraphics[height=4.5ex]{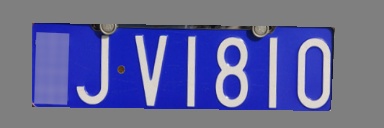}
		}
	}
			        
	\vspace{1.5mm}
			        
	\resizebox{0.85\linewidth}{!}{
		\subfloat[input]{
			\includegraphics[height=6.5ex]{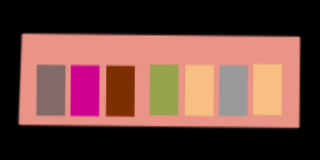}
			} \hspace{-3mm}
		\subfloat[output]{
			\includegraphics[height=6.5ex]{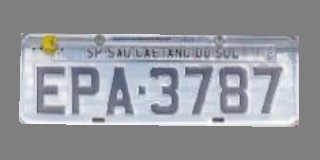}
			} \hspace{0.1mm}
		\subfloat[input]{
			\includegraphics[height=6.5ex]{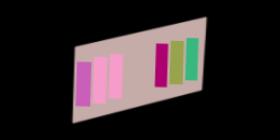}
			} \hspace{-3mm}
		\subfloat[output]{
			\includegraphics[height=6.5ex]{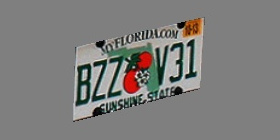}
			} \hspace{0.1mm}
		\subfloat[input]{
			\includegraphics[height=6.5ex]{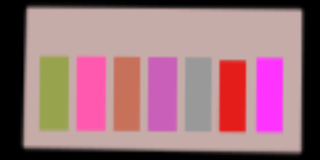}
			} \hspace{-3mm}
		\subfloat[output]{
			\includegraphics[height=6.5ex]{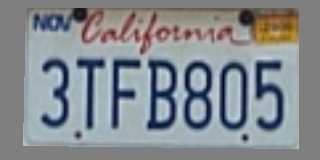}
		}
	}
			        
	\vspace{1.5mm}
			        
	\resizebox{0.85\linewidth}{!}{
		\subfloat[input]{
			\includegraphics[height=4.5ex]{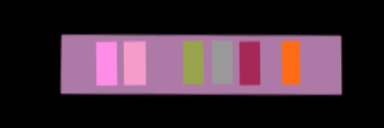}
			} \hspace{-3mm}
		\subfloat[output]{
			\includegraphics[height=4.5ex]{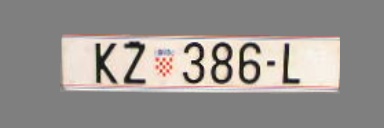}
			} \hspace{0.1mm}
		\subfloat[input]{
			\includegraphics[height=4.5ex]{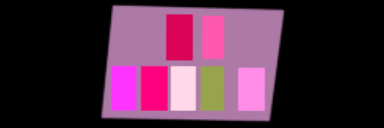}
			} \hspace{-3mm}
		\subfloat[output]{
			\includegraphics[height=4.5ex]{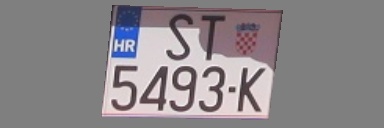}
			} \hspace{0.1mm}
		\subfloat[input]{
			\includegraphics[height=4.5ex]{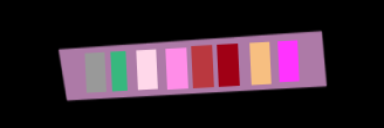}
			} \hspace{-3mm}
		\subfloat[output]{
			\includegraphics[height=4.5ex]{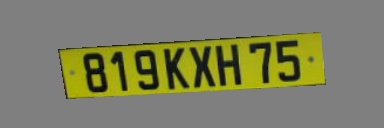}
		}
	}
	
	\vspace{-1.75mm}
			        
 \caption{Examples of image pairs used for training the pix2pix model. To create the input masks, labels are required for both the \gls*{lp}'s layout and corners, as well as for the bounding box of each~character.}
	\label{fig:samples-training-gan}
\end{figure}

The paired data required for training the pix2pix model \major{is} prepared as follows.
For each \gls*{lp} image generated through character permutation, which serves as the intended output, a corresponding segmentation mask \major{is} created to serve as the input.
These masks \major{are} designed such that each color represents a distinct \gls*{lp} layout class or character class.
For example, as shown in \cref{fig:samples-training-gan}, the digit `0' is indicated by a vivid red color (228, 28, 26), the letter `A' is denoted by a dark brown shade (126, 47, 0), the Mercosur layout is represented by a purplish-magenta tone (187, 0, 170), and the Chinese layout is denoted by a gray color (127, 127, 127).
The Glasbey library\customfootnote{\url{https://github.com/taketwo/glasbey}} was employed to generate a set of colors that \major{are} maximally distinguishable from each other.
Black (0, 0, 0) and similar shades \major{are} avoided in this process since black in the input mask represents the background.
Notably, the background in the output \gls*{lp} image consists of gray pixels.
This choice was made because using the original background led to inferior~results.

After completing the model's training, the next step involves using it to generate hundreds of thousands of new \gls*{lp} images.
Intuitively, this task \major{is} accomplished by feeding the model with segmentation masks derived from randomly selected \gls*{lp} layouts and character sequences.
While the characters \major{are} sampled from the valid alphabet per position, we \major{ensure} a balanced distribution of character classes at every position.

\begin{figure}[!htb]
        \centering
        
        \resizebox{0.8\linewidth}{!}{
        \includegraphics[width=0.18\linewidth]{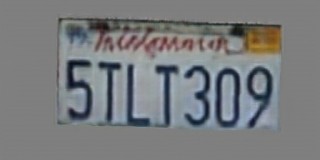} \hspace{-1.5mm}
        \includegraphics[width=0.18\linewidth]{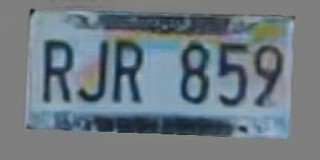} \hspace{-1.5mm}
        \includegraphics[width=0.18\linewidth]{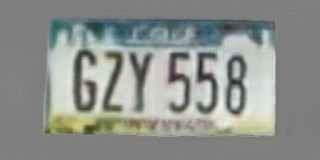} \hspace{-1.5mm}
        \includegraphics[width=0.18\linewidth]{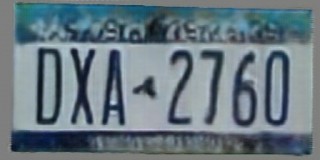} \hspace{-1.5mm}
        \includegraphics[width=0.18\linewidth]{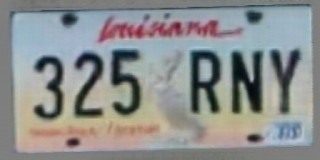}
        }
        
        \vspace{0.3mm}
        
        \resizebox{0.8\linewidth}{!}{
        \includegraphics[width=0.18\linewidth]{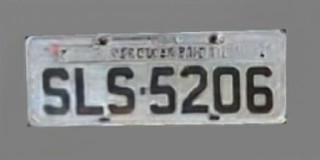} \hspace{-1.5mm}
        \includegraphics[width=0.18\linewidth]{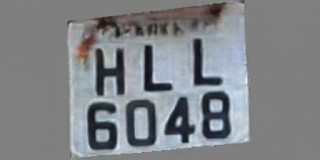} \hspace{-1.5mm}
        \includegraphics[width=0.18\linewidth]{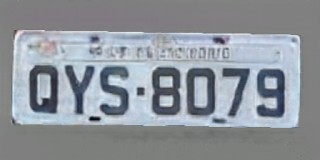} \hspace{-1.5mm}
        \includegraphics[width=0.18\linewidth]{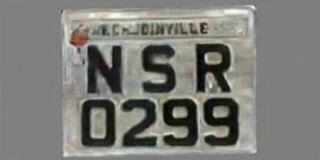} \hspace{-1.5mm}
        \includegraphics[width=0.18\linewidth]{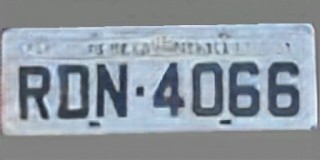}
        }
        
        \vspace{0.3mm}
        
        \resizebox{0.8\linewidth}{!}{
        \includegraphics[width=0.18\linewidth]{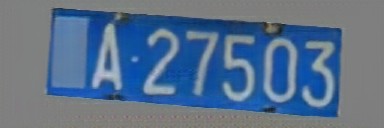} \hspace{-1.5mm}
        \includegraphics[width=0.18\linewidth]{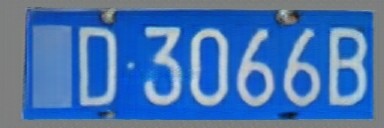} \hspace{-1.5mm}
        \includegraphics[width=0.18\linewidth]{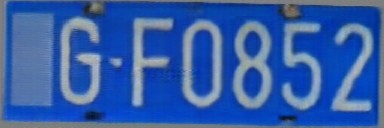} \hspace{-1.5mm}
        \includegraphics[width=0.18\linewidth]{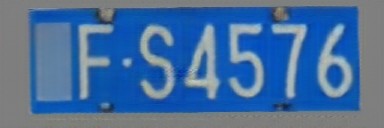} \hspace{-1.5mm}
        \includegraphics[width=0.18\linewidth]{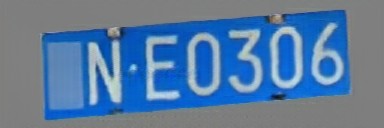}
        }
        
        \vspace{0.3mm}
        
        \resizebox{0.8\linewidth}{!}{
        \includegraphics[width=0.18\linewidth]{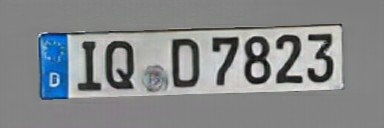} \hspace{-1.5mm}
        \includegraphics[width=0.18\linewidth]{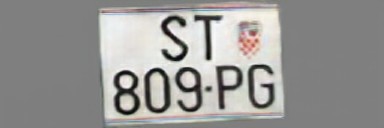} \hspace{-1.5mm}
        \includegraphics[width=0.18\linewidth]{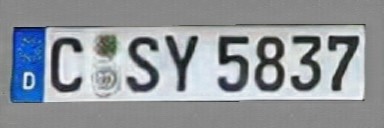} \hspace{-1.5mm}
        \includegraphics[width=0.18\linewidth]{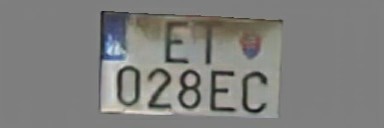} \hspace{-1.5mm}
        \includegraphics[width=0.18\linewidth]{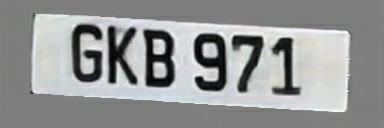}
        }
        
        \vspace{0.3mm}
        
        \resizebox{0.8\linewidth}{!}{
        \includegraphics[width=0.18\linewidth]{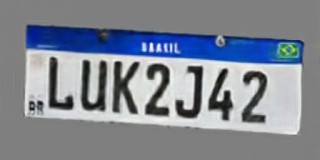} \hspace{-1.5mm}
        \includegraphics[width=0.18\linewidth]{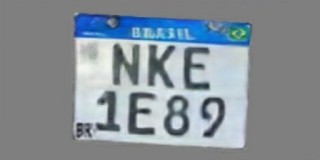} \hspace{-1.5mm}
        \includegraphics[width=0.18\linewidth]{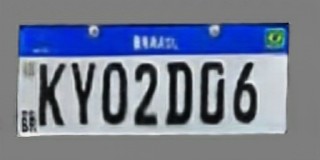} \hspace{-1.5mm}
        \includegraphics[width=0.18\linewidth]{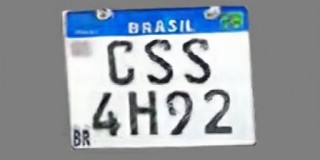} \hspace{-1.5mm}
        \includegraphics[width=0.18\linewidth]{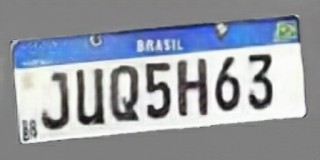}
        }
        
        \vspace{0.3mm}
        
        \resizebox{0.8\linewidth}{!}{
        \includegraphics[width=0.18\linewidth]{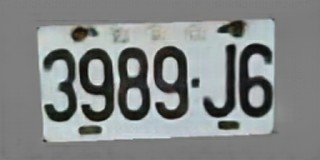} \hspace{-1.5mm}
        \includegraphics[width=0.18\linewidth]{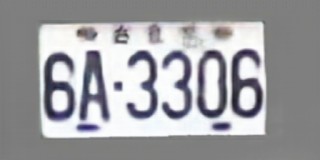} \hspace{-1.5mm}
        \includegraphics[width=0.18\linewidth]{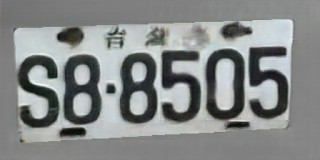} \hspace{-1.5mm}
        \includegraphics[width=0.18\linewidth]{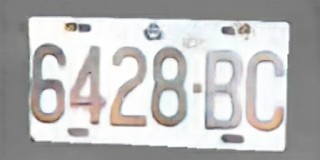} \hspace{-1.5mm}
        \includegraphics[width=0.18\linewidth]{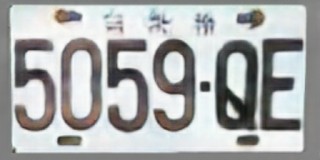}
        }

        \vspace{-2mm}

        \caption{Examples of selected images from those generated using pix2pix. From top to bottom, we show American, Brazilian, Chinese, European, Mercosur, and Taiwanese~\glspl*{lp}.}
        \label{fig:samples-gan-dataset}
    \end{figure}

Upon examining the generated \gls*{lp} images, we discovered that although many high-quality \glspl*{lp} were produced, a notable portion of them also displayed certain issues.
The primary issue identified was the distortion of characters or their blending into two distinct classes.
For instance, a generated character might exhibit a fusion of traits from `0' and `8', with the defining strokes that typically differentiate the two appearing faint and indistinct.
To address this matter, we ran the \fastocr model, which demonstrated superior cross-dataset results among a dozen recognition models in \citep{laroca2023leveraging}, on the millions of generated images and selected the top~$N$ predictions according to their associated confidence values.
Specifically, we selected the top $50$k images for each of the six explored \gls*{lp} layouts, totaling $300$k images.
This strategy proved effective in filtering out most images with defects, although it may have excluded some instances with a higher degree of variability.
Examples of the selected images are shown in~\cref{fig:samples-gan-dataset}.

It should be noted that we \major{train} the pix2pix model to produce a blurred representation instead of Chinese characters (this can be seen in \cref{fig:samples-training-gan,fig:samples-gan-dataset}).
This adjustment \major{is made} due to the absence of class labels for these characters in the training set.
Accurately labeling these characters poses a challenging task for individuals not proficient in Chinese, which is the case for our team. 
Further details on how we \major{handle} Chinese characters in our experiments can be found in \cref{sec:experiments-performance-evaluation}.

One might question the rationale behind employing segmentation maps as input for the \pixtwopix model, rather than using \gls*{lp} templates.
While we acknowledge that using templates as input would likely yield similar or even better results, the lack of \gls*{lp} style-related annotations in public datasets poses a challenge.
The provided information is limited to the geographical region where the images were collected (e.g., Europe, mainland China, and the United States).
Fundamentally, adopting \gls*{lp} templates as input would entail labeling the specific style of each \gls*{lp} and searching online platforms for the corresponding templates and character patches (or creating them using OpenCV or similar tools).
This is most likely why previous works explored very few \gls*{lp} styles in their experiments~\citep{zhang2021robust_attentional,fan2022improving,wang2022efficient}.

The major limitation of this method stems from its reliance on the training data, as it cannot synthesize \gls*{lp} layouts that are not included in the training set~\citep{gao2023group}.

\section{Experimental Setup}
\label{sec:experimental-setup}

This section describes the experimental setup adopted in this work. 
We first present the datasets explored, elucidating their division into training, validation and test subsets.
Subsequently, we list the \gls*{ocr} models we implemented for our assessments, providing the rationale for their choice over alternative options.
Lastly, we detail how the performance evaluation is~conducted.

While different machines were used for model training, all testing experiments were conducted on a PC equipped with an AMD Ryzen Threadripper $1920$X $3.5$GHz CPU, $96$~GB of RAM running at $2{,}133$ MHz, an SSD with read and write speeds of $3{,}500$~MB/s and $3{,}000$~MB/s respectively, and an NVIDIA Quadro RTX~$8000$ GPU~($48$~GB).

\subsection{Datasets}
\label{sec:experiments:datasets}

\begin{figure}[!htb]
    \centering
    \captionsetup[subfigure]{captionskip=2pt,justification=centering} 
    
    \resizebox{0.48\linewidth}{!}{
	\subfloat[][\caltech]{
    \includegraphics[height=6ex]{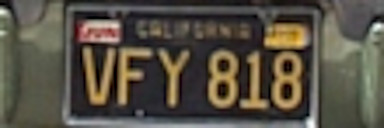}
    \includegraphics[height=6ex]{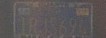}
    \includegraphics[height=6ex]{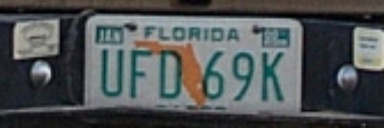}} 
    }
    \,
    \resizebox{0.48\linewidth}{!}{
	\subfloat[][\englishlp]{
    \includegraphics[height=6ex]{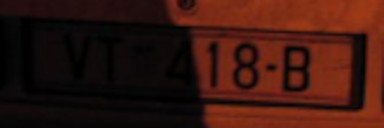}
    \includegraphics[height=6ex]{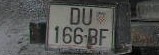}
    \includegraphics[height=6ex]{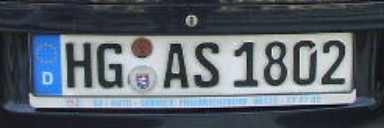}}  
    } \hspace{1mm}

    \vspace{1.25mm}
    
    \resizebox{0.48\linewidth}{!}{
	\subfloat[][\stills]{
    \includegraphics[height=6ex]{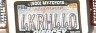}
    \includegraphics[height=6ex]{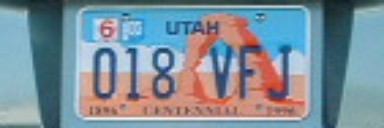}
    \includegraphics[height=6ex]{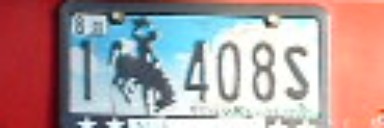}}
    }
    \,
    \resizebox{0.48\linewidth}{!}{
	\subfloat[][\chineselp\label{fig:samples-public-datasets-chineselp}]{
    \includegraphics[height=6ex]{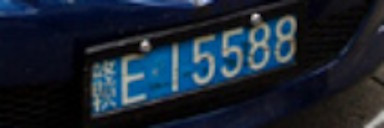}
    \includegraphics[height=6ex]{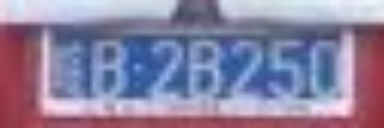}
    \includegraphics[height=6ex]{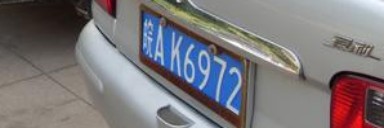}}  
    } \hspace{1mm}
    
    \vspace{1.25mm}
    
    \resizebox{0.48\linewidth}{!}{
	\subfloat[][\aolp]{
    \includegraphics[height=6ex]{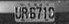}
    \includegraphics[height=6ex]{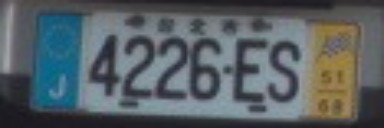}
    \includegraphics[height=6ex]{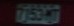}}
    }
    \,
    \resizebox{0.48\linewidth}{!}{
	\subfloat[][\openalpreu]{
    \includegraphics[height=6ex]{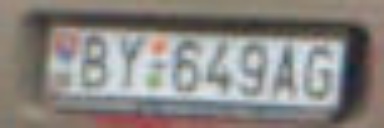}
    \includegraphics[height=6ex]{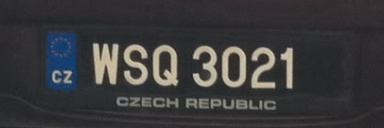}
    \includegraphics[height=6ex]{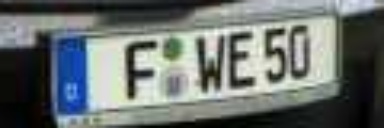}}  
    } \hspace{1mm}
    
    \vspace{1.25mm}
    
    \resizebox{0.48\linewidth}{!}{
	\subfloat[][\ssigsegplate]{
    \includegraphics[height=6ex]{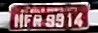}
    \includegraphics[height=6ex]{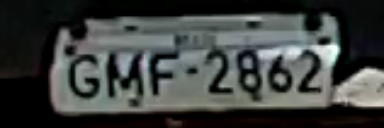}
    \includegraphics[height=6ex]{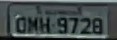}}
    }
    \,
    \resizebox{0.48\linewidth}{!}{
	\subfloat[][\pku]{
    \includegraphics[height=6ex]{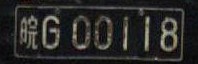}
    \includegraphics[height=6ex]{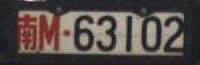}
    \includegraphics[height=6ex]{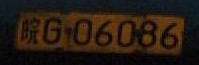}}
    } \hspace{1mm} 
    
    \vspace{1.25mm}
    
    \resizebox{0.48\linewidth}{!}{
	\subfloat[][\ufpralpr]{
    \includegraphics[height=6ex]{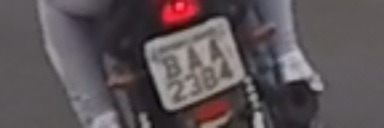}
    \includegraphics[height=6ex]{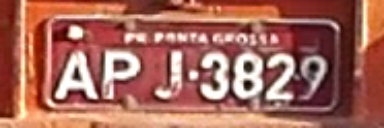}
    \includegraphics[height=6ex]{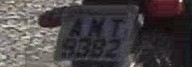}}
    }
    \,
    \resizebox{0.48\linewidth}{!}{
	\subfloat[][\cdhard]{
    \includegraphics[height=6ex]{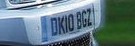}
    \includegraphics[height=6ex]{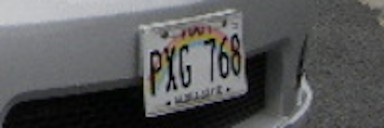}
    \includegraphics[height=6ex]{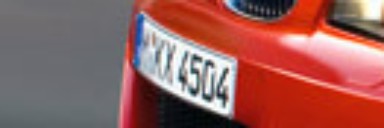}} 
    } \hspace{1mm} 

    \vspace{1.25mm}
    
    \resizebox{0.48\linewidth}{!}{
	\subfloat[][\clpd]{
    \includegraphics[height=6ex]{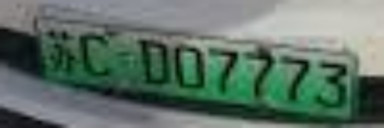}
    \includegraphics[height=6ex]{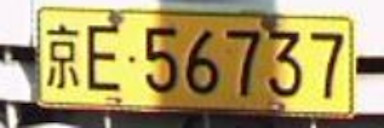}
    \includegraphics[height=6ex]{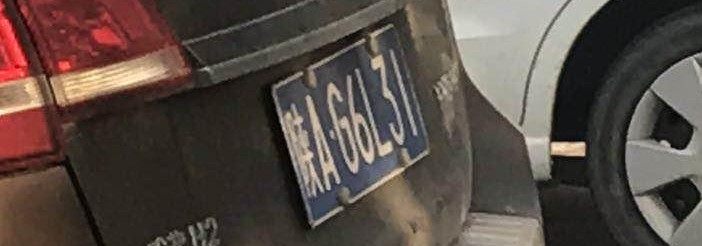}}
    }
    \,
    \resizebox{0.48\linewidth}{!}{
	\subfloat[][\rodosolalpr]{
    \includegraphics[height=5.9ex]{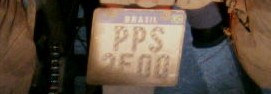}
    \includegraphics[height=5.9ex]{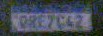}
    \includegraphics[height=5.9ex]{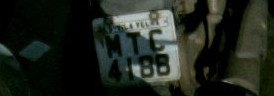}}
    } \hspace{1mm}

    \vspace{-2mm}
    
    \caption{\major{Representative \glspl*{lp} from the datasets used in our experiments, highlighting the diversity of \gls*{lp} layouts and visual characteristics. Adapted from~\citep{laroca2023leveraging}.}}
    \label{fig:samples-public-datasets}
\end{figure}

As shown in \cref{tab:experiments:overview_datasets}, our experiments were conducted on images from \numDatasets well-known public datasets gathered over the past two decades across distinct regions.
\major{\cref{fig:samples-public-datasets} presents representative \gls*{lp} images from these datasets, highlighting the variability in \gls*{lp} layouts and visual characteristics.}
\major{As detailed in the following paragraph, eight of these datasets were used for training, validation, and testing of the selected models, while the remaining four datasets were reserved exclusively for testing (cross-dataset~experiments).}

\begin{table}[!htb]
\centering
\renewcommand{\arraystretch}{1.05}
\caption{The $\numDatasets$ datasets used in our experiments.}
\label{tab:experiments:overview_datasets}

\vspace{0.3mm}

\resizebox{0.75\linewidth}{!}{ 
\begin{tabular}{@{}lccc@{}}
\toprule 
\textbf{Dataset} & \textbf{Images} & \textbf{Resolution} & \textbf{LP Layout} \\ \midrule
\caltech~\citep{caltech} & $126$ & $896\times592$ & American \\
\englishlp~\citep{englishlp} & $509$ & $640\times480$ & European \\
\stills~\citep{ucsd} & $291$ & $640\times480$ & American \\
\chineselp~\citep{zhou2012principal} & $411$ & Various & Chinese \\
\aolp~\citep{hsu2013application} & $2049$ & Various & Taiwanese \\
OpenALPR-EU$^\ast$~\citep{openalpr_eu} & $108$ & Various & European \\
\ssigsegplate~\citep{goncalves2016benchmark} & $2000$ & $1920\times1080$ & Brazilian \\
PKU$^\ast$~\citep{yuan2017robust} & $2253$ & $1082\times727$ & Chinese \\
\ufpralpr~\citep{laroca2018robust} & $4500$ & $1920\times1080$ & Brazilian \\
CD-HARD$^\ast$~\citep{silva2018license} & $102$ & Various & Various \\
CLPD$^\ast$~\citep{zhang2021robust_attentional} & $1200$ & Various & Chinese \\
\rodosolalpr~\citep{laroca2022cross} & $20000$ & $1280\times720$ & Brazilian \& Mercosur \\ \bottomrule \noalign{\vskip 0.2ex}
\multicolumn{4}{l}{\hspace{-2mm}$^\ast$\hspace{0.3mm}\small{Datasets used only for testing the deep models (i.e., cross-dataset experiments).}}
\end{tabular}
}
\end{table}

\major{For reproducibility, here we detail how we split the images from each dataset into training, validation and test sets\customfootnote{The complete list of which images from each dataset were used for training, validation and testing can be downloaded at \supplementarySplits}.
The \stills, \ssigsegplate, \ufpralpr and \rodosolalpr datasets were divided according to the protocols defined by the respective authors.
The other datasets, which do not have standard protocols, were split following prior studies.
More specifically, the images from the \openalpreu, \pku, \cdhard and \clpd datasets were used exclusively for testing, as in~\citep{silva2022flexible,wang2022rethinking,chen2023endtoend}.
The \caltech dataset was randomly divided into $63.5$\% of the images for training/validation and $36.5$\% for testing, as in~\citep{zhang2018vehicle,xiang2019lightweight,liu2024improving}.
Following~\cite{panahi2017accurate,henry2020multinational,beratoglu2021vehicle}, we randomly split the \englishlp dataset as follows: $80$\% of the images for training/validation and $20$\% for testing.
Regarding the \chineselp dataset, we followed the protocol adopted by \cite{laroca2021efficient,laroca2023leveraging}: $40$\% of the images for training, $20$\% for validation, and $40$\% for testing\customfootnote{To train the models, we excluded the few images from the \chineselp dataset that are also found in \clpd (both datasets include internet-sourced images, as discussed by~\cite{laroca2023do}).}.
Lastly, we divided each of the three subsets of the \aolp dataset (i.e., AC, LE, and RP) into training and test sets with a $2$:$1$ ratio, as~\cite{xie2018new,zhuang2018towards,liang2022egsanet}, and used $20$\% of the training images for~validation.}

To ensure a minimum of $500$ training images for each \gls*{lp} layout, we expanded our training set with $772$ images from the internet.
These images were labeled and made available by \cite{laroca2021efficient} and encompass $257$ American \glspl*{lp}, $347$ Chinese \glspl*{lp}, and $178$ European~\glspl*{lp}.
Furthermore, to address potential overfitting issues, we employed Albumentations~\citep{albumentations} --~a popular image augmentation library~-- to balance the number of training images from different datasets.

\major{We opted not to explore the \ccpd dataset~\citep{xu2018towards} in our experiments, despite its widespread use in the literature.
There are two primary reasons for this decision.
First, the dataset comprises highly compressed images, significantly reducing the legibility of the \glspl*{lp}~\citep{silva2022flexible}, and this does not align with our intended application.
\cite{qiao2021mango} even observed that some images within \ccpd are too blurry for the \glspl*{lp} to be recognized.
Second, the \ccpd dataset has experienced multiple updates and expansions since its introduction.
Consequently, there is an inconsistency regarding the dataset's size across different studies.
While some sources claim it contains $250$k images~\citep{liang2022egsanet,fan2022improving,ding2024endtoend}, others suggest a range of $280$-$290$k images~\citep{zou2020robust,wang2022rethinking,gao2023group}, whereas the current version has $366{,}789$ images.
The divergence in test sets across different versions renders the results reported in various studies not directly~comparable.}

\subsection{OCR Models}
\label{sec:experiments-methods}

This work compares $\numModels$ \gls*{ocr} models applied to the \gls*{lpr} task.
\cref{tab:models} presents an overview of these models, listing the original application for which they were designed and specifying the framework we used to implement them.

\setcounter{fndarknet}{\thefncounter}
\stepcounter{fncounter}
\setcounter{fnkeras}{\thefncounter}
\stepcounter{fncounter}
\setcounter{fnpytorch}{\thefncounter}
\stepcounter{fncounter}

\begin{table}[!htb]
\centering
\renewcommand{\arraystretch}{1.05}
\caption{The $\numModels$ \gls*{ocr} models explored in our experiments.}
\label{tab:models}

\vspace{0.35mm}

\resizebox{0.575\linewidth}{!}{%
\begin{tabular}{@{}llc@{}}
\toprule
\multicolumn{2}{c}{Model}                            & Original Application       \\ \midrule
\multicolumn{3}{l}{Framework: PyTorch\footnotemark[\thefnpytorch]}
\\
& \rtwoam~\citep{lee2016recursive}                             & Scene Text Recognition       \\
& \rare~\citep{shi2016robust}                            & Scene Text Recognition       \\
& \starnet~\citep{liu2016starnet}                         & Scene Text Recognition       \\
& \crnn~\citep{shi2017endtoend}                             & Scene Text Recognition       \\
& \grcnn~\citep{wang2017gated}                            & Scene Text Recognition       \\
& \rosetta~\citep{borisyuk2018rosetta}                          & Scene Text Recognition       \\
& \trba~\citep{baek2019what}                            & Scene Text Recognition       \\
& \vitstrbase~\citep{atienza2021vitstr}                      & Scene Text Recognition      \\
& \vitstrsmall~\citep{atienza2021vitstr}                      & Scene Text Recognition      \\
& \vitstrtiny~\citep{atienza2021vitstr}                      & Scene Text Recognition      \\ \midrule
\multicolumn{3}{l}{Framework: Keras\footnotemark[\thefnkeras]}
\\
& \holistic~\citep{spanhel2017holistic} & License Plate Recognition     \\ 
& \multitask~\citep{goncalves2018realtime}                        & License Plate Recognition      \\
& \multitaskLR~\citep{goncalves2019multitask}                        & License Plate Recognition      \\ 
& \cnng~\citep{fan2022improving}                        & License Plate Recognition      \\ \midrule
\multicolumn{3}{l}{Framework: Darknet\footnotemark[\thefndarknet]}
\\
& \crnet~\citep{silva2020realtime}                           & License Plate Recognition  \\   
& \fastocr~\citep{laroca2021towards}                        & Image-based Meter Reading               \\ \bottomrule
\end{tabular}%
}
\end{table}

We selected these models for two primary reasons.
First, they have a proven track record of success in \gls*{ocr} tasks (including but not limited to \gls*{lpr})~\citep{atienza2021vitstr,baek2019what,baek2021what,fan2022improving,silva2022flexible,nascimento2023super}.
Second, we are confident in our ability to train and adjust them effectively to ensure fairness in our experiments, as the authors provided enough details about the model architectures, and also because we designed/employed similar networks in previous works~\citep{goncalves2018realtime,goncalves2019multitask,laroca2021towards,laroca2023leveraging}. 
We are unaware of any work in the \gls*{alpr} literature where so many \gls*{ocr} models were explored in the~experiments.

Each model was trained and tested using either the framework in which it was originally implemented or well-known public repositories associated with it.
In summary, the YOLO-based models (i.e., \crnet and \fastocr) were implemented using the Darknet framework\customfootnote{\url{https://github.com/AlexeyAB/darknet/}}; the multi-task models (those listed in the middle section of \cref{tab:models}) were implemented using Keras\customfootnote{\url{https://keras.io/}}; and the other models were implemented using a popular fork of the open source repository of Clova AI Deep Text Recognition Benchmark\customfootnote{\url{https://github.com/roatienza/deep-text-recognition-benchmark/}}.

The hyperparameters used for training the models were defined based on preliminary experiments carried out in the validation set and are as follows.
In Darknet, we employed the \gls*{sgd} optimizer, $65$k iterations, batch size~=~$64$, and learning rate~=~[$10$\textsuperscript{-$3$},~$10$\textsuperscript{-$4$},~$10$\textsuperscript{-$5$}] with decay steps at $40$\% and $70$\% of the total iterations.
In Keras, we used the Adam optimizer, learning rate~=~$10$\textsuperscript{-$5$}, batch size~=~$64$, max epochs~=~$100$, and patience~=~$7$ (the number of epochs with no improvement after which training is stopped).
In PyTorch, we adopted the following parameters: Adadelta optimizer (decay rate $\rho=0.95$),  $300$k iterations, and batch size~=~$128$.

\subsection{Performance Evaluation}
\label{sec:experiments-performance-evaluation}

We present the models' performance for each dataset by calculating the ratio of correctly recognized \glspl*{lp} to the number of \glspl*{lp} in the test set.
We remark that an \gls*{lp} is deemed correctly recognized only if all its characters are precisely identified, given that a single misidentified character can lead to the misidentification of the~vehicle.

As mentioned in \cref{sec:introduction}, recent research has placed emphasis on the \gls*{lpr} stage~\citep{gao2023group,nascimento2023super,schirrmacher2023benchmarking,liu2024irregular}.
However, in our experiments, the \gls*{lp} patches fed into the recognition models were not cropped and rectified directly from the ground truth.
Instead, we detected the \glspl*{lp} in the original images using \yolocsp~\citep{wang2021scaledyolov4} and rectified them through a combination of \cdcc~\citep{laroca2021towards} --~for locating the \gls*{lp} corners~-- and perspective transformation.
We adopted this procedure to fairly compare our results with end-to-end \gls*{alpr} systems and to more accurately simulate real-world scenarios, where the \glspl*{lp} are not always optimally~detected.

\major{We chose \yolocsp over more recent models such as YOLOv10 and YOLOv11 for three practical and interconnected reasons. First, it is implemented in a fast and mature C/C++ version of the Darknet framework, supporting real-time processing in resource-constrained environments.
Second, it already delivers strong performance in our experiments, with an F-score of $98.5$\% at IoU~$>$~0.7 across all datasets (see \cref{tab:results-lp-detection} in \cref{sec:results:lp_detection_and_corner_detection}), which we consider sufficient for the purposes of this study. Third, its outputs are directly utilized by CDCC-NET to locate the LP corners used in the rectification step --~a crucial process that helps align \glspl*{lp} and reduce the impact of minor detection errors~(the rectification process is detailed in the next paragraph). Although we acknowledge that more recent YOLO versions generally offer higher detection accuracy, \yolocsp meets both the accuracy and efficiency requirements of our~pipeline.}

\begin{figure}[!htb]
    \centering
    \captionsetup[subfigure]{captionskip=1.75pt,justification=centering,font={small}}

    \resizebox{0.7\linewidth}{!}{
	\subfloat[][detected \gls*{lp} regions]{
    \includegraphics[width=0.2\linewidth]{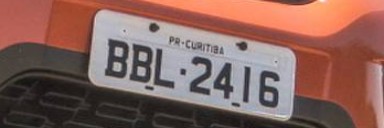}
    \includegraphics[width=0.2\linewidth]{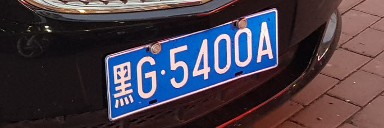}} \,

	\subfloat[][rectified \gls*{lp} regions]{
    \includegraphics[width=0.2\linewidth]{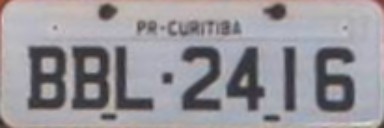}
    \includegraphics[width=0.2\linewidth]{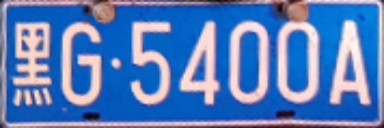}} \;
    }
    
    \vspace{-2.25mm}
    
    \caption{
    Two \glspl*{lp} before and after the rectification process.
    Observe that the rectified \glspl*{lp} resemble frontal views, becoming more horizontal, tightly bounded, and easier to~read.
    }
    \label{fig:rectification}
\end{figure}

We rectify each \gls*{lp} by calculating and applying a perspective transform from the coordinates of the four corners in the detected \gls*{lp} region to the corresponding vertices in the ``unwarped'' image.
These corresponding vertices were defined as follows: ($0$,~$0$) corresponds to the top-left corner; ($max_w-1$,~$0$) is the top-right corner; ($max_w-1$,~$max_h-1$) refers to the bottom-right corner; and ($0$,~$max_h-1$) indicates the bottom-left corner, where $max_w$ denotes the maximum distance between the top-right and top-left $x$ coordinates or the bottom-right and bottom-left $x$ coordinates, and $max_h$ is the maximum distance between the top-left and bottom-left $y$ coordinates or the top-right and bottom-right $y$ coordinates.
The rectification process is illustrated in \cref{fig:rectification}.
Recent works that exploited \gls*{lp} rectification to improve the recognition results include~\citep{fan2022improving,silva2022flexible,wang2022rethinking,xu2022eilpr,jiang2023efficient}.

It is essential to highlight that, in our experiments, we refrained from using prior knowledge about individual \gls*{lp} layouts to enhance the results through post-processing.
As an illustration, despite being aware that all \glspl*{lp} in a given dataset or particular region adhere to a fixed pattern (e.g., Brazilian \glspl*{lp} are composed of three letters followed by four digits), we treat the predictions made by the models as final.
We argue that by exposing the models to sufficient variability in the training stage, they can, to varying extents, implicitly learn and leverage such information to yield better~predictions.

We follow the methodology adopted by \cite{li2019toward} and \cite{laroca2022cross}, where all Chinese characters are collectively represented as a unified class denoted by `*'.
Accordingly, all results from other studies presented in our comparison with the state of the art (\cref{sec:results-cross_dataset}) were obtained in the same way, disregarding Chinese~characters.

\section{Results and Discussion}
\label{sec:results}

This section presents and analyzes the outcomes of our experiments.
\cref{sec:results:lp_detection_and_corner_detection} offers a concise overview of the results obtained in detecting the \glspl*{lp} and locating the corresponding corners.
The precise detection of the \gls*{lp} corners is pivotal for accurately rectifying the \glspl*{lp} before recognition.
\cref{sec:results:end-to-end} then delves into a detailed examination of the end-to-end results obtained by employing different \gls*{ocr}~models.

\subsection{LP Detection and Corner Detection}
\label{sec:results:lp_detection_and_corner_detection}

Various quantitative criteria can be employed to evaluate detection tasks.
Our assessment includes the widely adopted Precision, Recall and F-score metrics.
In line with prior studies~\citep{xu2018towards,jiang2023efficient,ke2023ultra}, we consider the detections correct when the \gls*{iou} with the ground truth exceeds~$0.7$.
Detections that meet this threshold typically encompass all \gls*{lp}~characters.

\cref{tab:results-lp-detection} presents the results obtained by \scaledyolo~\citep{wang2021scaledyolov4} and \iwpod~\citep{silva2022flexible} \major{(a well-known model specifically designed for \gls*{lp} detection)}.
Three key observations can be drawn from the results:
(i)~\scaledyolo demonstrated satisfactory results, both in terms of Precision and Recall, with instances of slightly lower Precision attributed to unlabeled \glspl*{lp} in the background of frames;
(ii)~while \iwpod directly predicts \gls*{lp} corners rather than bounding boxes, its performance is suboptimal in scenarios where the vehicles are far from the camera, as evidenced by the Recall rates reached in the \ufpralpr dataset;
and (iii)~\iwpod tends to predict a significant number of false positives, leading to notably low Precision rates.
Despite our exploration of higher detection thresholds, doing so led to the exclusion of many true \glspl*{lp} (lower Recall rates).
These observations likely influenced the decision of \cite{silva2022flexible} to feed regions identified by a vehicle detector (YOLOv3) into \iwpod instead of applying it directly to the original image (vehicle detection was not investigated in this study because most datasets lack labels for vehicle bounding boxes).
Balancing Precision and Recall is crucial for an efficient system operation, as it relies on accurately detecting all \glspl*{lp} while minimizing false~positives.

\begin{table}[!htb]
\centering
\setlength{\tabcolsep}{7pt}
\caption{Results obtained by \scaledyolo and \iwpod in the \gls*{lp} detection stage (@ \gls*{iou}~$>0.7$). For this evaluation, the corners predicted by \iwpod were converted into bounding boxes.}
\label{tab:results-lp-detection}

\vspace{0.5mm}

\resizebox{0.99\linewidth}{!}{%
\begin{tabular}{@{}ccccccccccc@{}}
\toprule
Model    & Metric & \multicolumn{1}{c}{\begin{tabular}[c]{@{}c@{}}\caltech\\\# $46$\end{tabular}} & \multicolumn{1}{c}{\begin{tabular}[c]{@{}c@{}}\englishlp\\\# $102$\end{tabular}} & \multicolumn{1}{c}{\begin{tabular}[c]{@{}c@{}}\stills\\\# $60$\end{tabular}} & \multicolumn{1}{c}{\begin{tabular}[c]{@{}c@{}}\chineselp\\\# $161$\end{tabular}} & \multicolumn{1}{c}{\begin{tabular}[c]{@{}c@{}}\aolp\\\# $687$\end{tabular}}    & \multicolumn{1}{c}{\begin{tabular}[c]{@{}c@{}}\ssigsegplate\\\# $804$\end{tabular}} & \multicolumn{1}{c}{\begin{tabular}[c]{@{}c@{}}\ufpralpr\\\# $1{,}800$\phantom{i}\end{tabular}} & \multicolumn{1}{c}{\begin{tabular}[c]{@{}c@{}}\rodosolalpr\\\# $8{,}000$\end{tabular}} & Average \\ \midrule
\scaledyolo & \multirow{2}{*}{Recall}        & $100.0$\%      & $\phantom{0}99.0$\%   & $100.0$\%     & $98.1$\%   & $99.9$\%      & $100.0$\%       & $99.2$\%       & $100.0$\%      & $\textbf{99.5}$\textbf{\%} \\
\iwpod &         & $\phantom{0}95.7$\%      & $100.0$\%   & $100.0$\%     & $97.5$\%   & $99.7$\%      & $\phantom{0}98.8$\%       & $82.4$\%       & $\phantom{0}99.6$\%      & $96.7$\% \\ \midrule
\scaledyolo & \multirow{2}{*}{Precision}        & $100.0$\%      & $\phantom{0}97.1$\%   & $\phantom{0}96.8$\%     & $98.1$\%   & $94.8$\%      & $\phantom{0}94.9$\%       & $97.8$\%       & $\phantom{0}99.6$\%      & $\textbf{97.4}$\% \\
\iwpod &         & $\phantom{0}66.7$\%      & $\phantom{0}77.9$\%   & $\phantom{0}73.2$\%     & $83.1$\%   & $88.3$\%      & $\phantom{0}61.6$\%       & $62.2$\%       & $\phantom{0}78.4$\%      & $73.9$\% \\ \midrule
\scaledyolo & \multirow{2}{*}{F-score}        & $100.0$\%      & $\phantom{0}98.1$\%   & $\phantom{0}98.4$\%     & $98.1$\%   & $97.3$\%      & $\phantom{0}97.5$\%       & $98.5$\%       & $\phantom{0}99.8$\%      & $\textbf{98.5}$\textbf{\%} \\
\iwpod &         & $\phantom{0}81.2$\%      & $\phantom{0}88.9$\%   & $\phantom{0}86.6$\%     & $90.3$\%   & $94.0$\%      & $\phantom{0}80.2$\%       & $72.3$\%       & $\phantom{0}89.0$\%      & $85.3$\% \\
\bottomrule
\end{tabular}%
}
\end{table}

To rectify the \glspl*{lp} found by \scaledyolo, it is necessary to locate the four corners associated with each of them.
\cref{tab:results-preprocessing-corner-detection} presents a comparison of the results obtained in this process by four models specifically designed for corner detection, including \iwpod.
The evaluation is carried out in terms of \lpnme~\citep{jia2023efficient}, a metric inspired by \gls*{nme}, which in turn is commonly employed to evaluate the quality of face alignment algorithms.
\lpnme is defined as follows:

\begin{small}
\begin{equation}
    \text{\textit{\lpnme}}(C, \hat{C}) = \frac{1}{4}\sum_{i=1}^{4}\frac{||C_{i} - \hat{C_{i}}||}{d} \; ,
\end{equation}
\end{small}

\noindent where $C$ and $\hat{C}$ are the ground truth and predicted corners, respectively, and $d$ is the normalization factor.
Following~\cite{jia2023efficient}, we adopt the diagonal length of the smallest bounding box that completely encloses the \gls*{lp} as the normalization~factor.

\begin{table}[!htb]
\centering
\setlength{\tabcolsep}{6pt}
\caption{Corner detection results achieved by four models within the regions found by \scaledyolo.
The results are presented in terms of \lpnme, where lower values indicate higher~accuracy.}
\label{tab:results-preprocessing-corner-detection}

\vspace{0.75mm}

\resizebox{0.99\linewidth}{!}{%
\begin{tabular}{@{}lccccccccc@{}}
\toprule
\diagbox[trim=l,innerrightsep=13.9pt]{Model}{Test set \# \glspl*{lp}}    & \multicolumn{1}{c}{\begin{tabular}[c]{@{}c@{}}\caltech\\\# $46$\end{tabular}} & \multicolumn{1}{c}{\begin{tabular}[c]{@{}c@{}}\englishlp\\\# $102$\end{tabular}} & \multicolumn{1}{c}{\begin{tabular}[c]{@{}c@{}}\stills\\\# $60$\end{tabular}} & \multicolumn{1}{c}{\begin{tabular}[c]{@{}c@{}}\chineselp\\\# $161$\end{tabular}} & \multicolumn{1}{c}{\begin{tabular}[c]{@{}c@{}}\aolp\\\# $687$\end{tabular}}    & \multicolumn{1}{c}{\begin{tabular}[c]{@{}c@{}}\ssigsegplate\\\# $804$\end{tabular}} & \multicolumn{1}{c}{\begin{tabular}[c]{@{}c@{}}\ufpralpr\\\# $1{,}800$\end{tabular}} & \multicolumn{1}{c}{\begin{tabular}[c]{@{}c@{}}\rodosolalpr\\\# $8{,}000$\end{tabular}} & Average \\ \midrule
\locatenet~\citep{meng2018robust}       & $0.0739$      & $0.0359$   & $0.0782$     & $0.1092$   & $0.0730$      & $0.0329$       & $0.0556$       & $0.0592$      & $0.0647$ \\
\hybridmobilenet~\citep{yoo2021deep} & $0.0323$      & $0.0226$   & $0.0352$     & $0.0391$   & $0.0332$       & $0.0214$       & $0.0313$       & $0.0383$      & $0.0317$ \\
\iwpod~\citep{silva2022flexible}       & $0.0244$      & $0.0143$   & $0.0205$     & $0.0138$   & $0.0205$      & $0.0098$       & $0.0194$       & $0.0141$      & $0.0171$ \\
\cdccnet~\citep{laroca2021towards}
        & $0.0160$      & $0.0117$   & $0.0164$     & $0.0176$   & $0.0142$       & $0.0098$       & $0.0168$       & $0.0150$      & $\textbf{0.0147}$ \\ \bottomrule

\end{tabular}%
}
\end{table}

\begin{figure}[!htb]
    \centering
    \captionsetup[subfigure]{labelformat=empty,captionskip=1.5pt,font=scriptsize}

    \subfloat[][]{\includegraphics[width=0.2\linewidth]{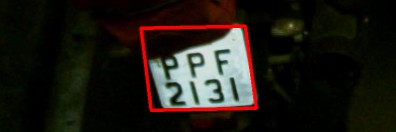}
    }
    \hspace{-1.5mm}
    \subfloat[][]{\includegraphics[width=0.2\linewidth]{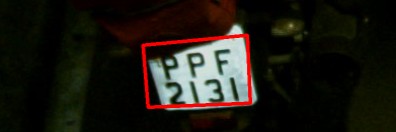}
    }
    \hspace{-1.5mm}
    \subfloat[][]{\includegraphics[width=0.2\linewidth]{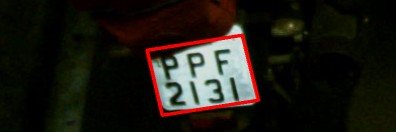}
    }
    \hspace{-1.5mm}
    \subfloat[][]{\includegraphics[width=0.2\linewidth]{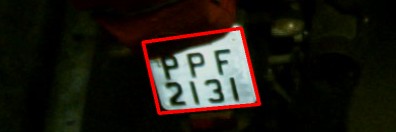}
    }

    \vspace{-3.1mm}

    \subfloat[][]{\includegraphics[width=0.2\linewidth]{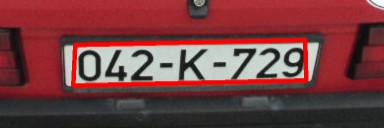}
    }
    \hspace{-1.5mm}
    \subfloat[][]{\includegraphics[width=0.2\linewidth]{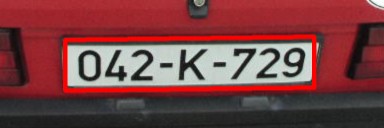}
    }
    \hspace{-1.5mm}
    \subfloat[][]{\includegraphics[width=0.2\linewidth]{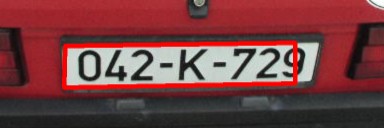}
    }
    \hspace{-1.5mm}
    \subfloat[][]{\includegraphics[width=0.2\linewidth]{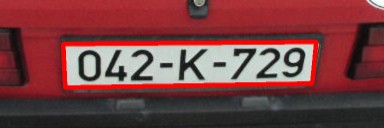}
    }

    \vspace{-3.1mm}

    \subfloat[][]{\includegraphics[width=0.2\linewidth]{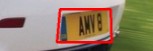}
    }
    \hspace{-1.5mm}
    \subfloat[][]{\includegraphics[width=0.2\linewidth]{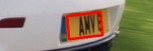}
    }
    \hspace{-1.5mm}
    \subfloat[][]{\includegraphics[width=0.2\linewidth]{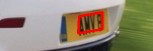}
    }
    \hspace{-1.5mm}
    \subfloat[][]{\includegraphics[width=0.2\linewidth]{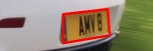}
    }

    \vspace{-3.1mm}

    \subfloat[][]{\includegraphics[width=0.2\linewidth]{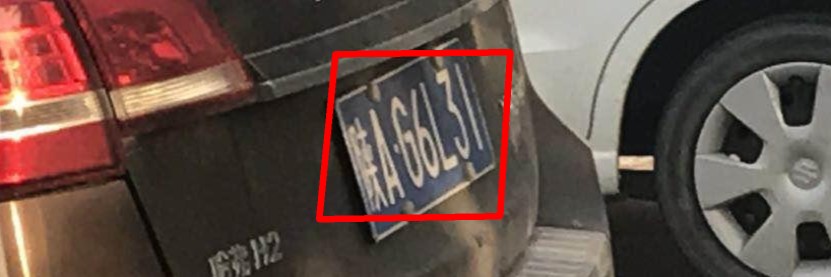}
    }
    \hspace{-1.5mm}
    \subfloat[][]{\includegraphics[width=0.2\linewidth]{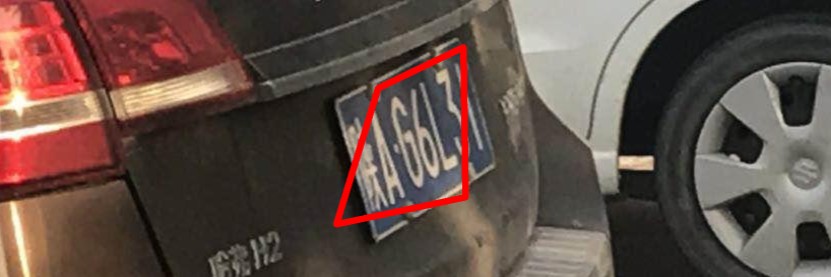}
    }
    \hspace{-1.5mm}
    \subfloat[][]{\includegraphics[width=0.2\linewidth]{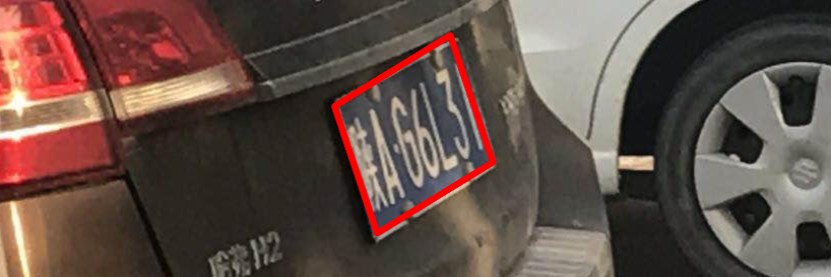}
    }
    \hspace{-1.5mm}
    \subfloat[][]{\includegraphics[width=0.2\linewidth]{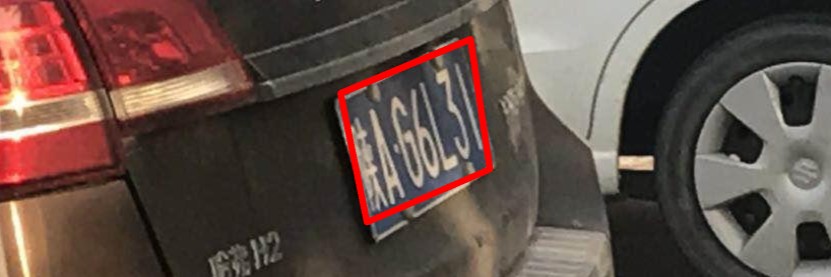}
    }

    \vspace{-3.1mm}

    \subfloat[][\locatenet]{\includegraphics[width=0.2\linewidth]{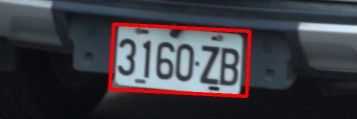}
    }
    \hspace{-1.5mm}
    \subfloat[][\centering \hybridmobilenet\hspace{0.1mm}]{\includegraphics[width=0.2\linewidth]{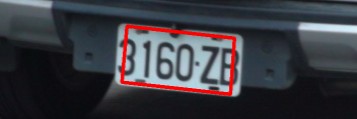}
    }
    \hspace{-1.5mm}
    \subfloat[][\iwpod]{\includegraphics[width=0.2\linewidth]{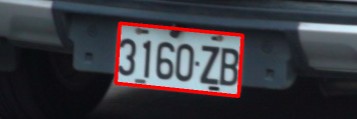}
    }
    \hspace{-1.5mm}
    \subfloat[][\cdcc]{\includegraphics[width=0.2\linewidth]{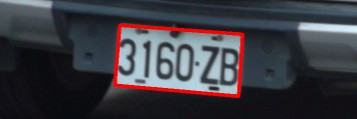}
    }

    \vspace{-2mm}
    
    \caption{Representative qualitative results achieved by four different models in corner detection.
    For better viewing, we draw a polygon from the predicted corner positions.}
    \label{fig:qualitative_results_preprocessing}
\end{figure}

\cdcc stands out as the top-performing model, achieving the lowest average \lpnme\ value of $0.0147$.
It is noteworthy, however, that the \iwpod model outperformed \cdcc in two datasets and achieved near-identical results in another.
\cref{fig:qualitative_results_preprocessing} showcases the predictions made by all models for five distinct \gls*{lp} images.
Although there is an evident similarity in the predictions for some \glspl*{lp}, the \cdcc model exhibits superior overall~accuracy.

The findings outlined in this section substantiate our choice to employ \yolocsp for \gls*{lp} detection and \cdcc for corner detection.
As elaborated in \cref{sec:experiments-performance-evaluation}, the corners predicted by \cdcc are used to rectify the \glspl*{lp} before~recognition.

\subsection{Overall Evaluation (End-To-End)}
\label{sec:results:end-to-end}

This section conducts a thorough comparative analysis of the \gls*{ocr} models, assessing their performance and contrasting the end-to-end results attained when employing the top-performing model with those reached by state-of-the-art approaches and established commercial systems (\cref{sec:results-intra_dataset,sec:results-cross_dataset,sec:results-state_of_the_art}).
Notably, the evaluation covers both intra- and cross-dataset scenarios.
Additionally, ablation studies are incorporated to demonstrate the impact of each explored method for generating synthetic images on the final results, as well as the importance of synthetic data when training data is scarce. 
Finally, \cref{sec:results:speed_accuracy_trade_off} examines the trade-off between speed and accuracy exhibited by the recognition models, highlighting those that strike a favorable~balance.

\subsubsection{Intra-Dataset Experiments}
\label{sec:results-intra_dataset}

\cref{tab:results-trad-rectified} presents the end-to-end results obtained across the disjoint test sets of the eight datasets used for training and validating the models.
In these experiments, all \gls*{ocr} models were trained using real images combined with synthetic ones generated by the three methods described in \cref{sec:synthetic}.
Later in this section, we present an ablation study that details the contribution of each image synthesis method to the results~achieved.

\begin{table}[!htb]
\centering
\renewcommand{\arraystretch}{1.05}
\setlength{\tabcolsep}{7pt}
\caption{Recognition rates obtained by all models under the intra-dataset protocol, where each model was trained once on the union of the training set images from these datasets (plus synthetic data) and evaluated on the respective test sets.
The best results achieved in each dataset are shown in~bold.}
\label{tab:results-trad-rectified}

\vspace{0.75mm}

\resizebox{0.99\textwidth}{!}{%
\begin{tabular}{@{}lccccccccc@{}}
\toprule
\diagbox[trim=l,innerrightsep=11.25pt]{Model}{Test set \# \glspl*{lp}}    & \multicolumn{1}{c}{\begin{tabular}[c]{@{}c@{}}\caltech\\\# $46$\phantom{i}\end{tabular}} & \multicolumn{1}{c}{\begin{tabular}[c]{@{}c@{}}\englishlp\\\# $102$\phantom{i}\end{tabular}} & \multicolumn{1}{c}{\begin{tabular}[c]{@{}c@{}}\stills\\\# $60$\phantom{\#}\end{tabular}} & \multicolumn{1}{c}{\begin{tabular}[c]{@{}c@{}}\chineselp\\\# $161$\phantom{i}\end{tabular}} & \multicolumn{1}{c}{\begin{tabular}[c]{@{}c@{}}\phantom{---}\aolp\phantom{---}\\\# $687$\phantom{i}\end{tabular}}  & \multicolumn{1}{c}{\begin{tabular}[c]{@{}c@{}}\ssigsegplate\\\# $804$\phantom{i}\end{tabular}} & \multicolumn{1}{c}{\begin{tabular}[c]{@{}c@{}}\ufpralpr\\\# $1{,}800$\phantom{i}\end{tabular}} & \multicolumn{1}{c}{\begin{tabular}[c]{@{}c@{}}\rodosolalpr\\\# $8{,}000$\phantom{i}\end{tabular}} & Average \\ \midrule
\cnng~\citep{fan2022improving} & $\textbf{97.8}$\textbf{\%}      & $91.2$\%   & $96.7$\%     & $98.8$\%   & $99.1$\%     & $98.8$\%       & $\textbf{96.1}$\textbf{\%}     & $97.1$\%      & $96.9$\% \\
\crnet~\citep{silva2020realtime}    & $93.5$\%      & $96.1$\%   & $98.3$\%     & $96.9$\%   & $98.7$\%     & $98.0$\%       & $89.3$\%     & \phantom{$^\dagger$}$88.3$\%$^\dagger$      & $94.9$\% \\
\crnn~\citep{shi2017endtoend}        & $93.5$\%      & $96.1$\%   & $96.7$\%     & $95.7$\%   & $98.8$\%      & $97.5$\%       & $87.0$\%       & $92.2$\%      & $94.7$\% \\
\fastocr~\citep{laroca2021towards}    & $95.7$\%      & $97.1$\%   & $95.0$\%     & $96.9$\%   & $98.7$\%     & $96.0$\%       & $89.6$\%       & \phantom{$^\dagger$}$88.1$\%$^\dagger$      & $94.6$\% \\
\grcnn~\citep{wang2017gated}       & $\textbf{97.8}$\textbf{\%}      & $\textbf{99.0}$\textbf{\%}   & $96.7$\%     & $98.8$\%   & $99.0$\%     & $97.9$\%       & $87.4$\%      & $93.0$\%      & $96.2$\% \\
\holistic~\citep{spanhel2017holistic}     & $95.7$\%      & $91.2$\%   & $93.3$\%     & $99.4$\%   & $99.3$\%     & $98.4$\%       & $94.9$\%       & $\textbf{97.9}$\textbf{\%}      & $96.3$\% \\
\multitask~\citep{goncalves2018realtime} & $\textbf{97.8}$\textbf{\%}      & $94.1$\%   & $\textbf{100.0}$\textbf{\%}     & $98.8$\%   & $99.1$\%      & $98.6$\%       & $93.3$\%       & $95.1$\%      & $97.1$\% \\
\multitaskLR~\citep{goncalves2019multitask}        & $95.7$\%      & $93.1$\%   & $93.3$\%     & $\textbf{100.0}$\textbf{\%}   & $99.6$\%      & $97.5$\%       & $94.6$\%       & $96.6$\%      & $96.3$\% \\
\rtwoam~\citep{lee2016recursive}        & $\textbf{97.8}$\textbf{\%}      & $94.1$\%   & $95.0$\%     & $98.8$\%   & $99.3$\%     & $99.3$\%       & $90.6$\%       & $94.4$\%      & $96.1$\% \\
\rare~\citep{shi2016robust}        & $\textbf{97.8}$\textbf{\%}      & $97.1$\%   & $98.3$\%     & $98.1$\%   & $99.4$\%     & $99.1$\%       & $91.9$\%       & $96.5$\%      & $97.3$\% \\
\rosetta~\citep{borisyuk2018rosetta}     & $95.7$\%      & $98.0$\%   & $98.3$\%     & $98.1$\%   & $98.7$\%     & $98.3$\%       & $92.6$\%       & $96.0$\%      & $97.0$\% \\
\starnet~\citep{liu2016starnet}    & $\textbf{97.8}$\textbf{\%}      & $\textbf{99.0}$\textbf{\%}   & $98.3$\%     & $98.1$\%   & $99.1$\%      & $99.3$\%       & $94.7$\%       & $97.0$\%      & $\textbf{97.9}$\textbf{\%} \\
\trba~\citep{baek2019what}        & $\textbf{97.8}$\textbf{\%}      & $\textbf{99.0}$\textbf{\%}   & $98.3$\%     & $98.8$\%   & $98.8$\%     & $99.3$\%       & $94.0$\%       & $97.3$\%     & $\textbf{97.9}$\textbf{\%} \\
\vitstrbase~\citep{atienza2021vitstr} & $95.7$\%      & $96.1$\%   & $93.3$\%     & $99.4$\%   & $\textbf{99.9}$\textbf{\%}      & $\textbf{99.4}$\textbf{\%}       & $94.6$\%       & $97.7$\%      & $97.0$\% \\ 
\vitstrsmall~\citep{atienza2021vitstr} & $95.7$\%      & $96.1$\%   & $98.3$\%     & $98.1$\%   & $99.1$\%      & $98.5$\%       & $94.9$\%       & $96.8$\%      & $97.2$\% \\ 
\vitstrtiny~\citep{atienza2021vitstr} & $93.5$\%      & $95.1$\%   & $91.7$\%     & $98.8$\%   & $99.0$\%      & $98.9$\%       & $92.3$\%       & $95.3$\%      & $95.5$\% \\ \midrule
Average & $96.2$\%      & $95.8$\%   & $96.4$\%     & $98.3$\%   & $99.1$\%      & $98.4$\%       & $92.4$\%       & $94.9$\%      & $\avgRectRR$\% \\ \bottomrule
\multicolumn{10}{l}{\rule{-3pt}{2.4ex}$^{\dagger}$\hspace{0.3mm}Images from the \rodosolalpr dataset were not used for training the \crnet and \fastocr models, as each character’s bounding box needs to be labeled for training them.}
\end{tabular}%
}
\end{table}

The first observation is that all models performed surprisingly well, reaching average recognition rates between $\minRR$\% and~$\maxRR$\%.
It is noteworthy that the mean results were well above $90$\% across all datasets, including \ufpralpr, which is known to be quite challenging~\citep{zhang2021vlpdr,zhou2023fafenet,ding2024endtoend}.
According to our analysis of the results (presented throughout this section), such impressive results are mainly due to the massive use of synthetic data combined with the \gls*{lp} rectification~stage.

Another point that immediately draws attention is that multiple models achieved the best result in at least one dataset.
For instance, the \cnng excelled in the \ufpralpr dataset, while the \multitaskLR and \holistic models reported the highest recognition rates on \chineselp and \rodosolalpr, respectively.
Interestingly, the models that performed better on average (i.e., \starnet and \trba) did not achieve the best results in six of the eight datasets;
some models actually reached the best result in one dataset and the worst in another (e.g., see the results achieved by the \cnng and \holistic models on the \englishlp dataset).
These results emphasize the importance of evaluating and comparing \gls*{ocr} models on various~datasets.

\begin{figure}[!htb]
	\centering
	\captionsetup[subfigure]{captionskip=1.5pt,labelformat=empty,font={footnotesize}}
	
	\resizebox{0.99\linewidth}{!}{
		\subfloat[][\centering \textbf{\texttt{\starnet}:}\,\;\texttt{ODB2B71}\hspace{\textwidth}\textbf{\texttt{\phantom{1111}\trba}:}\,\;\texttt{ODB2B71}]{
			\includegraphics[width=0.25\linewidth]{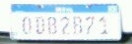}} \,
		\subfloat[][\centering \textbf{\texttt{\starnet}:}\,\;\texttt{DU166BF}\hspace{\textwidth}\textbf{\texttt{\phantom{-Net}\trba}:}\,\;\texttt{DU166BF}]{
			\includegraphics[width=0.25\linewidth]{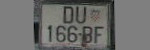}} \,
		\subfloat[][\centering \textbf{\texttt{\starnet}:}\,\;\texttt{HLP4594}\hspace{\textwidth}\textbf{\texttt{\phantom{-Net}\trba}:}\,\;\texttt{HLP4594}]{
			\includegraphics[width=0.25\linewidth]{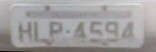}}		
             \,
		\subfloat[][\centering \textbf{\texttt{\starnet}:}\,\;\texttt{CKC3951}\hspace{\textwidth}\textbf{\texttt{\phantom{-Net}\trba}:}\,\;\texttt{CKC3951}]{
			\includegraphics[width=0.25\linewidth]{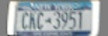}} \,
	}

	\vspace{3mm}

	\resizebox{0.99\linewidth}{!}{
		\subfloat[][\centering \textbf{\texttt{\starnet}:}\,\;\texttt{AWZ7648}\hspace{\textwidth}\textbf{\texttt{\phantom{-Net}\trba}:}\,\;\texttt{AWZ7648}]{
			\includegraphics[width=0.25\linewidth]{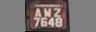}} \,
		\subfloat[][\centering \textbf{\texttt{\starnet}:}\,\;\texttt{*AS7603}\hspace{\textwidth}\textbf{\texttt{\phantom{-Net}\trba}:}\,\;\texttt{*AS7603}]{
			\includegraphics[width=0.25\linewidth]{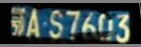}} \,
		\subfloat[][\centering \textbf{\texttt{\starnet}:}\,\;\texttt{PPR2D29}\hspace{\textwidth}\textbf{\texttt{\phantom{-Net}\trba}:}\,\;\texttt{PPR2D29}]{
			\includegraphics[width=0.25\linewidth]{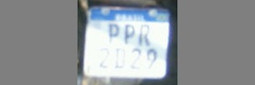}} \,
		\subfloat[][\centering \textbf{\texttt{\starnet}:}\,\;\texttt{WOBR3249}\hspace{\textwidth}\textbf{\texttt{\phantom{-Net}\trba}:}\,\;\texttt{WOBR3249}]{
			\includegraphics[width=0.25\linewidth]{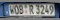}} \,
	}

        \vspace{3mm}
    
	\resizebox{0.99\linewidth}{!}{
		\subfloat[][\centering \textbf{\texttt{\starnet}:}\,\;\texttt{MR\red{D}3095}\hspace{\textwidth}\textbf{\texttt{\phantom{-Net}\trba}:}\,\;\texttt{MR\red{D}3095}]{
			\includegraphics[width=0.25\linewidth]{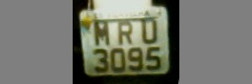}} \,
		\subfloat[][\centering \textbf{\texttt{\starnet}:}\,\;\texttt{\red{V}XS04R}\hspace{\textwidth}\textbf{\texttt{\phantom{-Net}\trba}:}\,\;\texttt{NXS04R}]{
			\includegraphics[width=0.25\linewidth]{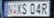}} \,
		\subfloat[][\centering \textbf{\texttt{\starnet}:}\,\;\texttt{*BD0D100}\hspace{\textwidth}\textbf{\texttt{\phantom{-Net}\trba}:}\,\;\texttt{*BD0\red{0}100}]{
			\includegraphics[width=0.25\linewidth]{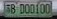}} \,
		\subfloat[][\centering \textbf{\texttt{\starnet}:}\,\;\texttt{LER\red{0}I79}\hspace{\textwidth}\textbf{\texttt{\phantom{-Net}\trba}:}\,\;\texttt{LERUI79}]{
			\includegraphics[width=0.25\linewidth]{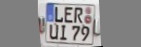}} \,
	}
	
	  \vspace{-1mm}
	    
	\caption{
        Predictions made for $12$ \gls*{lp} images by \starnet and \trba, the two models that exhibited the highest average performance in the intra-dataset experiments.
        Errors, if any, are highlighted in red.
        All \glspl*{lp} are well aligned because they were rectified before recognition, as detailed in \cref{sec:experiments-performance-evaluation}.
        }
	\label{fig:qualitative-results}
\end{figure}

\cref{fig:qualitative-results} showcases the predictions yielded by the \starnet and \trba models for \glspl*{lp} with distinct characteristics.
The outcomes underscore the models' robustness in handling diverse \gls*{lp} layouts, images with varying resolutions, \glspl*{lp} with different numbers of characters arranged in one or two rows, and scenarios where the characters are partially occluded.
Impressively, some of these \gls*{lp} styles were not even included in the training set.
Overall, errors are limited to instances where one character closely resembles another, often due to factors such as low resolution and artifacts on the \gls*{lp}.
Although this qualitative analysis focuses on the two models that achieved the best average results across the datasets, the other models generally produced similar~predictions.

\major{An important aspect to highlight is the effectiveness of synthetic data in scenarios with limited training samples --~a common issue in public datasets collected from specific regions.}
\major{\cref{tab:results-less-original-training-data} reports the average recognition rates of STAR-Net and TRBA when trained on progressively smaller subsets ($50$\%, $25$\%, $10$\%, $5$\% and $1$\%) of the original training set, comprising images from the eight datasets listed in \cref{sec:experiments:datasets}, both with and without the addition of synthetic data generated as described in \cref{sec:synthetic}.}
Remarkably, incorporating synthetic data in the training phase enabled commendable results to be reached even when using small fractions of the original training set.
For example, both \starnet and \trba achieved an average recognition rate exceeding $94.5$\% across all datasets when trained with only $10$\% of the original training set but supplemented with synthetic data.
In contrast, relying solely on real images with common transformations as data augmentation led to a substantial decline in the results.
Specifically, the recognition rates dropped below $75$\% when halving the original training set and plummeted to approximately $1$\% when using only $10$\% of it.
This underscores the effectiveness of synthetic data in mitigating the challenges posed by limited training~data.

\begin{table}[!htb]
\setlength{\tabcolsep}{7pt}
\centering
\caption{Average recognition rates obtained by \starnet and \trba when trained with reduced portions of the original training data. Naturally, images not included in the reduced training set were not used to generate synthetic images in the respective~experiments.}
\label{tab:results-less-original-training-data}

\vspace{0.75mm}

\resizebox{0.675\linewidth}{!}{
    \begin{tabular}{@{}lcccccc@{}}
    \toprule
    \diagbox[trim=l,innerrightsep=4.1pt]{\small{Model}}{\small{Real Images}}    & \phantom{-}$100$\%  & $50$\%   & $25$\%   & $10$\%   & $5$\%    & $1$\%    \\ \midrule
    \starnet (no synthetic) & \phantom{-}$95.3$\% & $62.0$\% & $18.3$\% & $\phantom{0}1.3$\% & $\phantom{0}0.2$\% & $\phantom{0}0.0$\% \\
    \starnet (w/ synthetic) & \phantom{-}$97.9$\% & $95.8$\% & $94.7$\% & $94.6$\% & $93.6$\% & $86.4$\% \\
    \noalign{\vskip 0.4ex}
    \cdashline{1-7} \noalign{\vskip 0.8ex}
    \trba (no synthetic)     & \phantom{-}$93.7$\% & $74.0$\% & $23.9$\% & $\phantom{0}0.9$\% & $\phantom{0}0.2$\% & $\phantom{0}0.0$\% \\
    \trba (w/ synthetic)     & \phantom{-}$97.9$\% & $97.0$\% & $96.0$\% & $94.5$\% & $94.3$\% & $87.9$\% \\ \bottomrule
    \end{tabular}
}
\end{table}

\major{\cref{tab:ablation} elucidates the effectiveness of each image synthesis method described in \cref{sec:synthetic}, as well as their combination, to the results obtained.
It reveals that each method contributes considerably to enhancing the results.
Notably, a substantial synergistic effect is observed when combining these methods, pushing the performance boundaries of recognition models applied to \gls*{lpr}. 
To elaborate, the best recognition rates (i.e., $\avgRR$\% and $\avgRectRR$\% for unrectified and rectified \glspl*{lp}, respectively), on average for all models, were achieved by combining original data with images synthesized in all three ways.
To further support these findings, paired t-tests confirmed statistically significant differences~($p < 0.01$) between the top-performing configurations in both setups.}

When real images were combined solely with images generated through character permutation, as in~\citep{shashirangana2022license,laroca2023leveraging}, the average recognition rates obtained were $\avgOrigPermutation$\% and $\avgOrigPermutationRect$\% for unrectified and rectified \glspl*{lp}, respectively.
Combining real images with \gls*{lp} templates alone, as in~\citep{maier2022reliability,gao2023group}, resulted in average recognition rates of $\avgOrigTemplates$\% and $\avgOrigTemplatesRect$\% for unrectified and rectified \glspl*{lp}, respectively.
Finally, the combination of real images with those generated through a \gls*{gan} model (in our case, pix2pix), as in~\citep{zhang2021robust_attentional,shvai2023multiple}, yielded average recognition rates of $\avgOrigGAN$\% and $\avgOrigGANRect$\% for unrectified and rectified \glspl*{lp}, respectively.
\major{Note that training with \gls*{lp} templates alone, without any real images, resulted in much lower performance, with average recognition rates below $50$\%, likely due to the significant domain difference between these synthetic images and real~\glspl*{lp}.}

\begin{table}[!htb]
\centering
\renewcommand{\arraystretch}{1.025}
\caption{
Average recognition rates \major{and corresponding standard deviations} obtained across all models and datasets with different types of images included in the training set.
\major{The values shown below each image synthesis method represent the number of images used for that method.
``Data aug.'' refers to online data augmentation (using standard transformations) applied to real images.}
The synergistic impact of the three image synthesis methods in enhancing the overall results is evident.
\major{As creating synthetic images through character permutation and \gls*{gan} relies on the existence of real images, their integration is evaluated only in scenarios where real images are included in the training~set.}
}
\label{tab:ablation}

\vspace{0.75mm}

\resizebox{0.7\linewidth}{!}{%
    \begin{tabular}{@{}cccccc@{}}
        \toprule
        \begin{tabular}[c]{@{}c@{}}Real Images\\15k (+ data aug)\end{tabular} & \begin{tabular}[c]{@{}c@{}}Templates\\100k\end{tabular} & \begin{tabular}[c]{@{}c@{}}Permutation\\300k\end{tabular} & \begin{tabular}[c]{@{}c@{}}\gls*{gan}\\300k\end{tabular}    & Average & Average (rect.)  %

\\ \midrule
                 & \cmark             &     &        & $42.5$\% $\pm$ $13.9$\% & $46.5$\% $\pm$ $11.5$\% \\
                 
                \cmark   &             &           &        & $84.5$\% $\pm$ $10.5$\% & $88.1$\% $\pm$ $\phantom{0}8.1$\% \\
        \cmark   &       &     \cmark      &        & $\avgOrigPermutation$\% $\pm$ $\phantom{0}3.6$\% & $\avgOrigPermutationRect$\% $\pm$ $\phantom{0}2.3$\% \\
        \cmark   &     \cmark        &     &        & $\avgOrigTemplates$\% $\pm$ $\phantom{0}2.7$\% & $\avgOrigTemplatesRect$\% $\pm$ $\phantom{0}1.3$\% \\
        \cmark   &             &           & \cmark & $\avgOrigGAN$\% $\pm$ $\phantom{0}2.1$\% & $\avgOrigGANRect$\% $\pm$ $\phantom{0}1.5$\% \\
        \cmark   & \cmark      & \cmark    &        & $93.8$\% $\pm$ $\phantom{0}1.6$\% & $95.5$\% $\pm$ $\phantom{0}0.8$\% \\
        \cmark   &       &     \cmark      & \cmark & $94.0$\% $\pm$ $\phantom{0}1.6$\% & $95.6$\% $\pm$ $\phantom{0}1.2$\% \\
        \cmark   &    \cmark         &     & \cmark & $94.1$\% $\pm$ $\phantom{0}1.7$\% & $95.8$\% $\pm$ $\phantom{0}1.0$\% \\
        \cmark   & \cmark      & \cmark    & \cmark & $\textbf{\avgRR}$\textbf{\%} $\pm$ \phantom{i}$\textbf{1.6}$\textbf{\%}\phantom{i} & $\textbf{\avgRectRR}$\textbf{\%} $\pm$ \phantom{i}$\textbf{1.1}$\textbf{\%}\phantom{i} \\ \bottomrule
    \end{tabular}%
}

\end{table}

It is important to highlight how much better the results were when training the models with both real and synthetic images (i.e., $\avgRR$\% and $\avgRectRR$\%) compared to those obtained when simply training the models with original images augmented by common transformations such as random rotation, random noise, random cropping, random compression, and random changes in brightness, saturation, and contrast (i.e., $84.5$\% and~$88.1$\%).

It is also noteworthy that both the templates and the images produced by the \gls*{gan} model contributed significantly more to improving the \gls*{ocr} models' performance than the images generated through character permutation.
This observation aligns with the fact that the images created via character permutation share many characteristics with their original counterparts (e.g., character position, compression artifacts, camera noise, among others) despite having different character~sequences.

\subsubsection{Cross-Dataset Experiments}
\label{sec:results-cross_dataset}

\major{As the performance of \gls*{lp} recognition under the traditionally adopted intra-dataset protocol --~where models are trained and tested on disjoint subsets of the same dataset~-- is rapidly improving, many researchers argue that a more realistic assessment comes from cross-dataset experiments.
Such experiments better reflect real deployments where new cameras are deployed more frequently than models are retrained~\citep{laroca2022cross,laroca2022first,wang2022rethinking,schirrmacher2023benchmarking}.
Accordingly, \cref{tab:results-cross} presents the results obtained by the same models used for the intra-dataset evaluation~(\cref{tab:results-trad-rectified}), now tested on four entirely unseen datasets.
No adjustments were made to the models, ensuring a faithful representation of their cross-dataset~performance.}

\begin{table}[!htb]
\centering
\renewcommand{\arraystretch}{1.05}
\setlength{\tabcolsep}{8pt}
\caption{Recognition rates obtained by all models on four public datasets that were not seen during the training stage (cross-dataset experiments).
The best results for each dataset are shown in~bold.}

\label{tab:results-cross}

\vspace{0.75mm}

\resizebox{0.8\linewidth}{!}{%
\begin{tabular}{@{}lccccc@{}}
\toprule
\diagbox[trim=l,innerrightsep=11.75pt]{Model}{Dataset \# \glspl*{lp}}   & \multicolumn{1}{c}{\begin{tabular}[c]{@{}c@{}}\openalpreu\\\# $108$\phantom{i}\end{tabular}} & \multicolumn{1}{c}{\begin{tabular}[c]{@{}c@{}}\pku\\\# $2{,}253$\phantom{i}\end{tabular}} & {\begin{tabular}[c]{@{}c@{}}\cdhard\\\# $104$\phantom{i}\end{tabular}} & \multicolumn{1}{c}{\begin{tabular}[c]{@{}c@{}}\clpd\\\# $1{,}200$\phantom{i}\end{tabular}} & Average \\ \midrule
\cnng~\citep{fan2022improving} & $95.4$\% & $98.6$\% & $58.7$\% & $92.9$\% & $86.4$\% \\
\crnet~\citep{silva2020realtime} & $93.5$\% & $\textbf{99.5}$\textbf{\%} & $67.3$\%  & $92.9$\%   & $88.3$\% \\
\crnn~\citep{shi2017endtoend} & $97.2$\% & $99.1$\% & $56.7$\%       & $94.2$\%   & $86.8$\% \\
\fastocr~\citep{laroca2021towards} & $98.1$\% & $99.1$\% & $69.2$\%      & $94.4$\%   & $90.2$\% \\
\grcnn~\citep{wang2017gated} & $97.2$\%   & $99.0$\% & $57.7$\% &  $94.5$\%   & $87.1$\% \\
\holistic~\citep{spanhel2017holistic} & $95.4$\% & $99.0$\% & $54.8$\%        & $94.0$\%   & $85.8$\% \\
\multitask~\citep{goncalves2018realtime}            & $96.3$\%        & $98.8$\% & $54.8$\%  & $93.7$\% & $85.9$\% \\
\multitaskLR~\citep{goncalves2019multitask}            & $94.4$\%        & $98.8$\% & $53.8$\%  & $92.6$\% & $84.9$\% \\
\rtwoam~\citep{lee2016recursive}          & $98.1$\%        & $99.4$\% & $57.7$\%  & $93.8$\% & $87.3$\% \\
\rare~\citep{shi2016robust} & $\textbf{99.1}$\textbf{\%} & $99.1$\% & $72.1$\%     & $95.2$\%   & $91.4$\% \\
\rosetta~\citep{borisyuk2018rosetta}          & $97.2$\%       & $99.2$\% & $64.4$\%   & $93.8$\% & $88.7$\% \\
\starnet~\citep{liu2016starnet}       & $98.1$\%       & $98.5$\%  & $71.2$\% & $95.0$\% & $90.7$\% \\
\trba~\citep{baek2019what} & $\textbf{99.1}$\textbf{\%} & $99.4$\% & $\textbf{76.9}$\textbf{\%}       & $\textbf{96.2}$\textbf{\%}   & $\textbf{92.9}$\textbf{\%}  \\
\vitstrbase~\citep{atienza2021vitstr} & $94.4$\% & $99.0$\% & $54.8$\%        & $93.4$\%   & $85.4$\% \\ 
\vitstrsmall~\citep{atienza2021vitstr} & $96.3$\% & $97.4$\%  & $59.6$\%      & $94.3$\%   & $86.9$\% \\
\vitstrtiny~\citep{atienza2021vitstr} & $94.4$\% & $97.6$\% & $53.8$\%        & $92.3$\%   & $84.5$\% \\ \midrule
Average & $96.5$\% & $98.8$\%    & $61.5$\%    & $93.9$\%   & $87.7$\% \\ \bottomrule
\end{tabular}%
}
\end{table}

These results demonstrate that the explored \gls*{ocr} models, trained on a combination of real and synthetic images, maintain high performance even in unseen scenarios.
What most caught our attention was the consistency of the \trba model~\citep{baek2019what}, as it also reached the best results in this evaluation.
On the other hand, here the \starnet model (which tied with the best results in the intra-dataset experiments) was outperformed by \rare in all datasets.
\major{Based on these findings, we consider the configuration combining \emph{\yolocsp} for detection, \emph{\cdcc} for rectification, and \emph{\trba} for recognition as the most effective setup in our benchmark. 
Accordingly, we adopt this configuration for comparisons with state-of-the-art approaches in the next~section.}

While subpar results were achieved on the \cdhard dataset, it is essential to recognize the inherent complexity of this dataset, as implied by its name.
Our analysis has revealed that the primary challenge posed by this dataset lies in the diverse range of \gls*{lp} layouts it encompasses.
Images within the dataset feature vehicles from various regions not represented in the datasets used for model training, such as Dubai and New South Wales.
The high degree of tilt of many \glspl*{lp} would further hinder recognition if not rectified before the recognition~stage~\citep{silva2018license,silva2022flexible}.

\subsubsection{Comparison With Previous Works and Commercial Systems}
\label{sec:results-state_of_the_art}

\major{In \cref{tab:results-sota}, we compare the end-to-end results obtained by the best-performing model combination in our benchmark with those reported by state-of-the-art \gls*{alpr} systems.}
Following common practice, to make the comparison fair, we only consider systems evaluated in the same way as in our benchmark (see details in \cref{sec:experiments:datasets}).
We also compare our results with those obtained by \cite{sighthoundapi} and \cite{openalprapi}, which are two commercial systems widely adopted as baselines in the literature~\citep{weihong2020research,lubna2021automatic}.

\begin{table}[!htb]
\centering
\renewcommand{\arraystretch}{1.05}
\setlength{\tabcolsep}{7pt}
\caption{
Recognition rates obtained by our best approach (which uses \trba as the recognition model), state-of-the-art methods, and two commercial systems in the eight datasets where part of the images was used for training the networks.
The best results achieved in each dataset are shown in~bold.
}
\label{tab:results-sota}

\vspace{0.75mm}

\resizebox{0.99\textwidth}{!}{%
\begin{tabular}{@{}lccccccccc@{}}
\toprule
\diagbox[trim=l,innerrightsep=19.5pt]{Approach}{Test set}    & \multicolumn{1}{c}{\begin{tabular}[c]{@{}c@{}}\caltech\\\# $46$\phantom{i}\end{tabular}} & \multicolumn{1}{c}{\begin{tabular}[c]{@{}c@{}}\englishlp\\\# $102$\phantom{i}\end{tabular}} & \multicolumn{1}{c}{\begin{tabular}[c]{@{}c@{}}\stills\\\# $60$\phantom{\#}\end{tabular}} & \multicolumn{1}{c}{\begin{tabular}[c]{@{}c@{}}\chineselp\\\# $161$\phantom{i}\end{tabular}} & \multicolumn{1}{c}{\begin{tabular}[c]{@{}c@{}}\phantom{---}\aolp\phantom{---}\\\# $687$\phantom{i}\end{tabular}}  & \multicolumn{1}{c}{\begin{tabular}[c]{@{}c@{}}\ssigsegplate\\\# $804$\phantom{i}\end{tabular}} & \multicolumn{1}{c}{\begin{tabular}[c]{@{}c@{}}\ufpralpr\\\# $1{,}800$\phantom{i}\end{tabular}} & \multicolumn{1}{c}{\begin{tabular}[c]{@{}c@{}}\rodosolalpr\\\# $8{,}000$\phantom{i}\end{tabular}} & Average \\ \midrule
\cite{sighthoundapi} & $87.0$\%      & $93.1$\%   & $96.7$\%     & $95.0$\%   & $95.5$\%     & $82.8$\%       & $62.9$\%     & $57.0$\%      & $83.7$\% \\
\cite{castrozunti2020license}$^{\ddagger}$  & $91.3$\%      & $-$   & $\textbf{98.3}$\textbf{\%}     & $-$   & $-$     & $-$       & $-$     & $-$      & $-$ \\
\cite{silva2022flexible} & $-$      & $-$   & $-$     & $-$   & $97.4$\%     & $-$       & $86.3$\%     & $-$     & $-$ \\
\cite{henry2020multinational} & $\textbf{97.8}$\textbf{\%}      & $97.1$\%   & $-$     & $-$   & $98.9$\%     & $-$       & $-$     & $-$      & $-$ \\
\cite{laroca2021efficient} (run $1$)$^{\dagger}$ & $\textbf{97.8}$\textbf{\%}      & $96.1$\%   & $96.7$\%     & $98.1$\%   & $\textbf{99.4}$\textbf{\%}     & $98.8$\%       & $89.7$\%     & $-$      & $-$ \\
\cite{zhou2023fafenet} & $-$      & $-$   & $-$     & $-$   & $-$     & $-$       & $90.3$\%     & $-$     & $-$ \\

\cite{silva2022flexible}$^{\dagger}$ & $-$      & $-$   & $-$     & $-$   & $99.0$\%     & $-$       & $91.8$\%     & $-$     & $-$ \\
\cite{laroca2023leveraging} & $87.0$\%      & $88.2$\%   & $86.7$\%     & $96.9$\%   & $\textbf{99.4}$\textbf{\%}      & $95.8$\%       & $89.7$\%       & $95.6$\%      & $92.4$\% \\ 
\cite{openalprapi}$^{\dagger}$ & $95.7$\%      & $98.0$\%   & $\textbf{98.3}$\textbf{\%}     & $96.9$\%   & $97.1$\%     & $93.0$\%       & $92.2$\%     & $69.3$\%     & $92.6$\% \\
\cite{chen2023endtoend} & $-$      & $-$   & $-$     & $-$   & $-$      & $-$       & $-$       & $96.6$\%      & $-$ \\ 
\cite{nascimento2023super}$^{\ddagger}$ & $-$      & $-$   & $-$     & $-$   & $-$      & $-$       & $-$       & $96.6$\%      & $-$ \\ 
Ours & $87.0$\%      & $91.2$\%   & $88.3$\%     & $98.1$\%   & $98.4$\%      & $98.1$\%       & $92.1$\%       & $96.8$\%      & $93.7$\% \\ 
\cite{zhang2021vlpdr} & $-$      & $-$   & $-$     & $-$   & $-$     & $98.6$\%       & $92.3$\%     & $-$     & $-$ \\
\cite{liu2024improving}$^{\ddagger}$ & $-$      & $-$   & $-$     & $-$   & $99.0$\%      & $-$       & $-$       & $97.0$\%      & $-$ \\
\textbf{Ours + synthetic} & $\textbf{97.8}$\textbf{\%}      & $\textbf{99.0}$\textbf{\%}   & $\textbf{98.3}$\textbf{\%}     & $\textbf{98.8}$\textbf{\%}   & $98.8$\%      & $\textbf{99.3}$\textbf{\%}       & $\textbf{94.0}$\textbf{\%}       & $\textbf{97.3}$\textbf{\%}      & $\textbf{97.9}$\textbf{\%} \\ \bottomrule
\multicolumn{10}{l}{\rule{-3pt}{2.4ex}$^{\dagger}$\hspace{0.3mm}\gls*{alpr} systems that rely on pre-defined heuristic rules (prior knowledge) to refine the predictions returned by the \gls*{ocr}~model.} \\
\multicolumn{10}{l}{\rule{-3pt}{2.4ex}$^{\ddagger}$\hspace{0.3mm}The \gls*{lp} patches fed into the \gls*{ocr} model were cropped directly from the ground truth in~\citep{castrozunti2020license,nascimento2023super,liu2024improving}.}
\end{tabular}%
}
\end{table}

\major{It is impressive that, without using any heuristic rules or post-processing, the benchmark's top-performing setup (using \trba for recognition) achieves state-of-the-art performance on all datasets except \aolp.
Note that higher recognition rates (e.g., $99.9$\%) were actually attained on the \aolp dataset using other \gls*{ocr} models (see \cref{tab:results-trad-rectified}); however, we do not consider those results here because the respective models did not outperform \trba on average across the full~benchmark.}

Two other aspects should be highlighted from the above results. 
First, the positive influence of exploiting synthetic data is reaffirmed, as our system did not achieve the best results on most datasets when solely using real data (plus simple data augmentation) for training.
Second, both the \cite{sighthoundapi} and \cite{openalprapi} commercial systems performed poorly on the \rodosolalpr dataset (with $57.0$\% and $69.3$\% recognition rates, respectively).
As shown in \cref{tab:commercial-systems-rodosol}, the primary reason for such underwhelming results is that these systems do not work well for motorcycle \glspl*{lp} (which are challenging in nature, having two rows of characters and being smaller in size) or Mercosur \glspl*{lp} (which were adopted just a few years ago). 
These observations underscore the importance of comparing \gls*{alpr} systems across diverse datasets that encompass various collection methodologies, feature images of different types of vehicles (including motorcycles), and exhibit different \gls*{lp} layouts (including two-row configurations).

\begin{table}[!htb]
\centering
\caption{Results achieved by two well-known commercial systems in the \rodosolalpr dataset.
It can be seen that their capabilities vary considerably according to the vehicle type and the \gls*{lp}~layout.
}
\label{tab:commercial-systems-rodosol}

\vspace{0.75mm}

\resizebox{0.55\linewidth}{!}{
\begin{tabular}{@{}ccccc@{}}
\toprule
\multirow{2}{*}{System} & \multicolumn{2}{c}{Vehicle Type} & \multicolumn{2}{c}{LP Layout} \\ 
                        & Cars         & Motorcycles       & Brazilian      & Mercosur     \\ \midrule
\cite{sighthoundapi}              & 81.3\%       & 32.7\%            & 63.9\%         & 50.1\%       \\
\cite{openalprapi}                & 95.6\%       & 43.0\%            & 90.7\%         & 47.8\%       \\ \bottomrule
\end{tabular}
}
\end{table}

There are many recent works where the authors evaluated the generalizability of the proposed methods in the \pku~\citep{yuan2017robust} and \clpd~\citep{zhang2021robust_attentional} datasets, both collected in mainland China.
\major{Hence, in \cref{tab:results-pku-clpd}, we compare the results obtained by these methods (plus Sighthound and OpenALPR) with those reached by the best-performing configuration in our benchmark.}
\major{For each approach, we also report the number of real Chinese \glspl*{lp} used during training and indicate whether the method qualifies as multinational, defined here as not being trained or fine-tuned exclusively on Chinese~LPs.}

\begin{table}[!htb]
\centering
\caption{%
Comparison of the recognition rates (\%) obtained by our best approach (TRBA), state-of-the-art methods, and commercial systems on the \clpd and \pku datasets.
These experiments assess the generalizability of these \gls*{alpr} approaches, as no images from those datasets were used for training.
The methods categorized as ``Multinational'' were not trained or fine-tuned exclusively on Chinese~\glspl*{lp}.
\major{Additional experiments show that our pipeline further outperforms others even when trained using the \ccpd dataset, highlighting its effectiveness in both low-resource and large-scale~scenarios.}
}
\label{tab:results-pku-clpd}

\vspace{0.75mm}

\resizebox{0.95\linewidth}{!}{%
\begin{tabular}{@{}lcccc@{}}
\toprule
\multirow{2}{*}{Approach}                                         & \multirow{2}{*}{\begin{tabular}[c]{@{}c@{}}Real images of Chinese\\\glspl*{lp} used for training\end{tabular}} & \multirow{2}{*}{Multinational} & \multicolumn{2}{c}{Recognition Rate} \\
& & & \clpd & \pku \\ \midrule
\cite{sighthoundapi} & ?\phantom{\plus}                                                                               & \cmark                                                                                    & $85.2$\% & $89.3$\%                                                    \\
\cite{zhang2021robust_attentional} & $100{,}000$\plus                                                                          & \phantom{\red{\xmark}}                                                                                    & $87.6$\% & $90.5$\%                                                   \\
\cite{fan2022improving}                    & $100{,}000$\plus                                                                          & \cmark                                                                                    & $88.5$\% & $92.5$\%                                                    \\
Ours          & $506$\phantom{\plus}                                                                             & \cmark                                                                                    & $90.1$\% & $96.8$\%                                           \\
\cite{rao2024license}$^\dagger$ & $4{,}444$\phantom{\plus}                                                                           & \phantom{\red{\xmark}}                                                                                    & $91.4$\% & $96.1$\% \\
\cite{liu2021fast}                                & $10{,}000$\phantom{\plus}                                                                           & \phantom{\red{\xmark}}                                                                                    & $91.7$\% & $-$                                                    \\
\cite{openalprapi}                                  & ?\phantom{\plus}                                                                               & \phantom{\red{\xmark}}                                                                                    & $91.8$\% & $96.0$\%                                                     \\
\cite{chen2023endtoend}                    & $100{,}000$\plus                                                                          & \phantom{\red{\xmark}}                                                                                    & $92.4$\% & $92.8$\% \\
\cite{ke2023ultra} & $100{,}000$\plus                                                                          & \phantom{\red{\xmark}}                                                                                    & $93.2$\% & $-$ \\
\cite{zou2020robust}                                & $100{,}000$\plus                                                                          & \phantom{\red{\xmark}}                                                                                    & $94.0$\% & $96.6$\%                                                     \\
\cite{zou2022license}                                & $100{,}000$\plus                                                                          & \phantom{\red{\xmark}}                                                                                    & $94.5$\% & $-$                                                     \\
\cite{wang2022efficient}                   & $100{,}000$\plus                                                                          & \phantom{\red{\xmark}}                                                                                    & $94.8$\% & $-$ \\ 
\cite{wang2022rethinking}                    & $100{,}000$\plus                                                                          & \phantom{\red{\xmark}}                                                                                    & $95.3$\% & $96.9$\% \\ 
\textbf{Ours + synthetic}          & $\textbf{506}$\phantom{\plus}                                                                             & \textbf{\cmark}                                                                                    & $\textbf{96.2}$\textbf{\%} & $\textbf{99.4}$\textbf{\%}                                            \\[0.4ex] \cdashline{1-5} \noalign{\vskip 0.8ex} 
[Additional experiments] & & & & \\
\hspace{2mm}Ours + \ccpd's training set         & $100{,}000$\plus                                                                    & \cmark                                                                                    & $94.5$\% & $96.8$\%                                           \\
\hspace{1.8mm}\textbf{Ours + \ccpd's training set + synthetic}          & $\textbf{100{,}000}$\textbf{\plus}                                                                            & \textbf{\cmark}                                                                                    & $\textbf{\OursCCPDCLPD}$\textbf{\%} & $\textbf{\OursCCPDPKU}$\textbf{\%}                                          \\
\bottomrule
\multicolumn{5}{l}{\rule{-3pt}{2.1ex}\footnotesize$^{\dagger}$\hspace{0.3mm}Approaches in which we applied the authors' code and pre-trained models to obtain the reported results.} \\
\end{tabular}%
}
\end{table}

\major{When exploring synthetic data for training the \gls*{ocr} model, the end-to-end approach (\yolocsp + \cdcc + \trba) exhibited significantly superior performance compared to state-of-the-art methods and commercial systems on both datasets.}
These results are particularly noteworthy given that our training dataset comprised only~$506$ real images of vehicles with Chinese \glspl*{lp}, while most baseline models were trained on over $100{,}000$ images from the \ccpd dataset~\citep{xu2018towards}.
\major{Indeed, this is one of the reasons why this configuration did not outperform the baselines even further, especially on the \clpd dataset, as several of the recognition errors occurred on \gls*{lp} styles missing in our training set but present in \ccpd (e.g., 8-character green \glspl*{lp} from electric vehicles).}
\major{To assess the upper-bound performance in a comparable setup, we conducted an additional experiment by incorporating LP images from CCPD's training set into the training data, consistent with previous studies.
In this setting, the benchmark configuration achieved recognition rates of {\OursCCPDCLPD}\% on \clpd and {\OursCCPDPKU}\% on \pku, demonstrating its strong generalization ability when trained with a more comprehensive~dataset.}

\subsubsection{Speed/Accuracy Trade-Off}
\label{sec:results:speed_accuracy_trade_off}

\begin{figure}[!htb]
    \centering

    \includegraphics[width=0.725\linewidth]{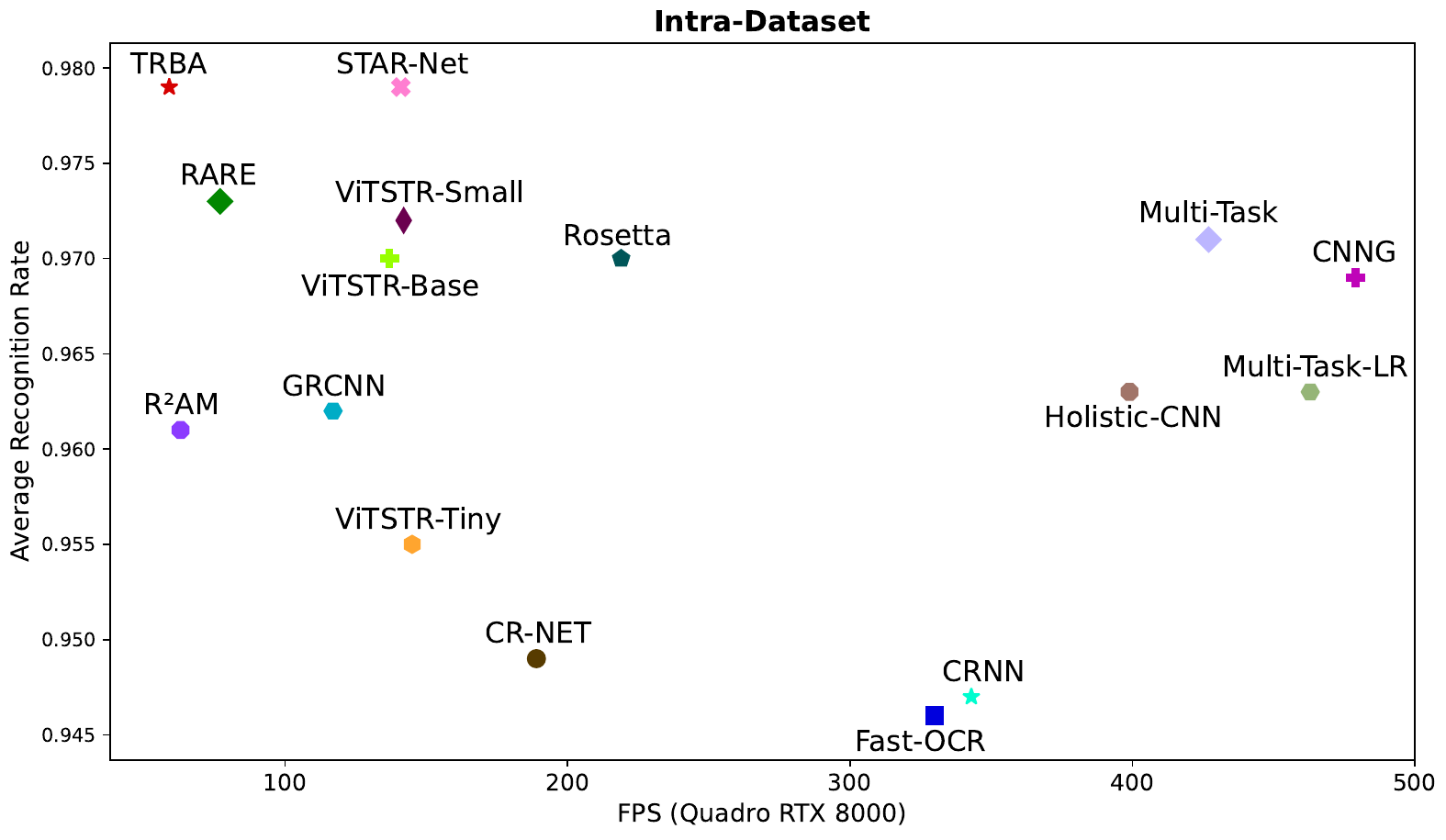}
    
    \vspace{2mm}
    
     \includegraphics[width=0.725\linewidth]{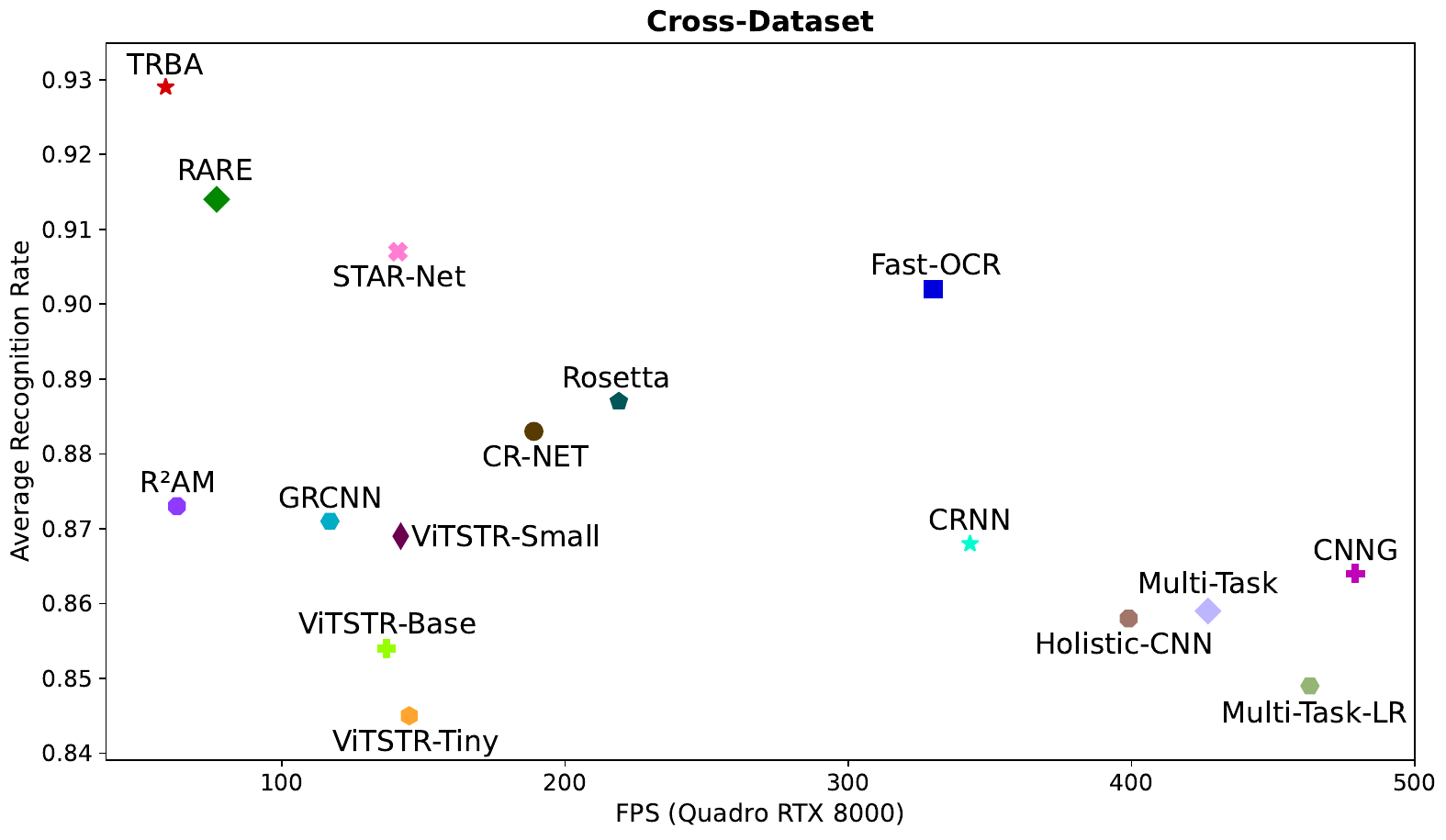}
    
    \vspace{-3mm}

    \caption{Average recognition rate across datasets and the corresponding \gls*{fps} processing capabilities for all \gls*{ocr} models on intra-dataset~(top) and cross-dataset~(bottom) experiments. The specific \gls*{fps} value for each model is as follows: \cnng:~$479$; \crnet:~$189$; \crnn:~$343$; \fastocr:~$330$; \grcnn:~$117$; \holistic:~$399$; \multitask:~$427$; \multitaskLR:~$463$; \rtwoam:~$63$; \rare:~$77$; \rosetta:~$219$; \starnet:~$141$; \trba:~$59$; \vitstrbase:~$137$; \vitstrsmall:~$142$; and \vitstrtiny:~$145$.}
    \label{fig:trade-off-recognition-rate-speed-fps}
\end{figure}

The importance of devising methods that strike an optimal balance between speed and accuracy has been highlighted in recent \gls*{alpr} research~\citep{jiang2023efficient,ke2023ultra,ding2024endtoend}.
Thus, this section examines the speed/accuracy trade-off of the \gls*{ocr} models explored in this study.
\cref{fig:trade-off-recognition-rate-speed-fps} compares the average recognition rates reached across datasets and the corresponding \gls*{fps} processing capabilities of all models, both in intra- and cross-dataset~setups.

In \emph{intra-dataset} scenarios, the multi-task models, particularly \multitask and \cnng, demonstrated an exceptional balance between speed and accuracy.
This can be attributed to their ability to learn potential classes for each character position independently, thereby avoiding confusion between similar letters and digits in layouts where they appear in distinct positions.
When the primary goal is to achieve the utmost recognition rate across various scenarios, \starnet is a more compelling option than \trba.
This is due to \starnet achieving the same average recognition rate as \trba ($97.9$\%) while processing more than twice the \gls*{fps} ($141$ vs.~$59$).

In \emph{cross-dataset} scenarios, as outlined in \cref{sec:results-cross_dataset}, \trba once again emerged as the top performer in terms of average recognition rate, standing alone this time, while \starnet was outperformed by \rare.
Concerning the trade-off between speed and accuracy, the \fastocr model clearly excels, striking a commendable balance between the two.
Its relatively high accuracy on unseen \glspl*{lp} can be attributed to its foundation on the YOLO object detector.
Consequently, it detects and recognizes each character individually, as opposed to predicting specific \gls*{lp} sequences that mimic patterns from the training set.
Conversely, the multi-task models experienced a substantial decline in recognition rate precisely because they learned to predict sequences based on patterns observed in the training set, which often differ from those observed in other datasets/scenarios.

Regarding the \vitstr variants, it is worth noting that they handle essentially the same number of \gls*{fps}.
This is because the key differentiation among the \vitstrbase, -Small and -Tiny models lies in their respective number of parameters and computations required (FLOPS), rather than in the number of \gls*{fps} they can process~\citep{atienza2021vitstr}.

\section{Conclusions}
\label{sec:conclusions}

This paper delves into the integration of real and synthetic data for improved \acrfull*{lpr}.
Synthetic \gls*{lp} images were generated using three widely adopted methodologies in the literature: a rendering-based pipeline (templates), character permutation, and a \gls*{gan} model.
We subjected $\numModels$ \gls*{ocr} models to a thorough benchmarking process involving \numDatasets public datasets acquired from various regions.
The experiments encompassed both intra- and cross-dataset evaluations, including an examination of the speed/accuracy trade-off of the models.
\major{To the best of our knowledge, this constitutes the most extensive experimental evaluation conducted in the~field.}

Several key findings \major{emerge} from our study.
Primarily, the massive use of synthetic data significantly \major{improves} the performance of all models, both in intra- and cross-dataset scenarios.
The quantitative and qualitative results demonstrated the models' robustness in effectively handling diverse \gls*{lp} layouts, images with varying resolutions, and \glspl*{lp} with varying numbers of characters arranged in one or two rows.
Notably, employing the top-performing \gls*{ocr} model~(\trba) yielded end-to-end results that surpassed those reached by state-of-the-art methods and established commercial systems.
These results are particularly noteworthy as our models were not specifically trained for each \gls*{lp} layout, and we refrained from incorporating heuristic rules to enhance the predictions for \glspl*{lp} from specific regions through post-processing --~a departure from many existing methods.
This streamlined approach significantly simplifies the process of incorporating support for \glspl*{lp} from new regions or even markedly different \gls*{lp} styles within the same~region.

The conducted ablation studies provide two important insights.
\major{First, each synthesis method contributes considerably to enhancing the results, and a substantial synergistic effect is observed when they were combined.
This finding contrasts with the common practice of generating synthetic \glspl*{lp} exclusively through a single methodology.
Second, incorporating synthetic data into the training set enables commendable results to be attained even when using small fractions of the original data.
This highlights the effectiveness of synthetic data in overcoming the challenges posed by scarce training~data.}

Acknowledging the significance of both model speed and accuracy in real-world applications, we investigated how well the models strike a balance between these two factors.
Although the multi-task models demonstrated an impressive speed/accuracy trade-off in intra-dataset scenarios, this optimal balance did not extend to cross-scenario scenarios.
In such instances, these models exhibited a more substantial decline in recognition rates than most other models. 
Remarkably, in cross-dataset scenarios, \fastocr stood out due to its great balance between speed and accuracy.
The effectiveness of \fastocr in cross-dataset scenarios can be attributed to its character-level detection and recognition approach, setting it apart from other models that predict \gls*{lp} sequences by replicating patterns from the training set.
While this replication approach proves effective in similar contexts, its efficacy tends to diminish when applied to different regions or~scenarios.

\major{An additional noteworthy contribution of this work lies in our commitment to granting access to all synthetic images generated for training the models.
We will also provide the corresponding code for generating new \gls*{lp} images using each synthesis~method.}

It is essential to acknowledge the extensive number of experiments conducted for this study.
We carried out \numTimesTrainedWords training sessions for each of the $\numModels$ recognition models under investigation (refer to \cref{tab:ablation}), subjecting them to testing across various seen and unseen datasets.
We also explored the \pixtwopix model's capabilities for generating \gls*{lp} images and performed multiple experiments related to the \gls*{lp} detection and corner detection tasks, as reported in \cref{tab:results-lp-detection,tab:results-preprocessing-corner-detection}.
Note that a single training process for some models (e.g., \trba and \vitstrbase) took several days to complete on an NVIDIA Quadro RTX $8000$ GPU, which is currently one of the top-performing GPUs in the~market.

\subsection{Future Directions}

\major{Despite the comprehensive scope of our study and the significantly greater dataset diversity compared to prior work, we recognize an important area for improvement regarding the geographic, script, and layout variability of \glspl*{lp}.
In particular, our evaluation did not include datasets with non-Latin scripts, such as Arabic or Cyrillic, nor \glspl*{lp} with radically different formats, such as vertical or non-standard layouts.
These omissions may affect the global applicability of our findings and highlight the need for future research to assess synthetic data generation methods and \gls*{ocr} models in regions with alternative scripts and diverse \gls*{lp}~layouts.}

\major{Building on the demonstrated effectiveness of synthetic data for accurately recognizing \glspl*{lp} in high-quality images captured under diverse conditions and across multiple regions, we advocate for a gradual shift in the focus of ALPR research.
Specifically, there is a pressing need to address the challenges posed by low-resolution or low-quality \glspl*{lp}.
These difficult scenarios --~often encountered in criminal investigations where the \gls*{lp} remains unclear across all video frames~-- remain significantly underexplored in the current~literature.}

\major{Future work should also consider pairing our synthetic data pipeline with architecture-level innovations, such as neural architecture search or lightweight adapters, to further improve recognition performance, especially in unconstrained~environments.}

\major{Although synthetic data is widely regarded as a privacy-preserving solution, we acknowledge that its use in ALPR systems does not eliminate broader ethical concerns, such as the potential for misuse in large-scale surveillance or data manipulation. In addition to providing an ethical use guideline in the public code repository accompanying this work, we encourage future research to investigate methods that improve traceability and accountability in synthetic data pipelines. In particular, exploring mechanisms such as data watermarking and auditing tools is essential to support responsible use and reduce the risk of misuse in sensitive surveillance contexts.}

\section*{Acknowledgments}

This work was supported by the Coordination for the Improvement of Higher Education Personnel~(CAPES) and the National Council for Scientific and Technological Development~(CNPq) (grant number~315409/2023-1).
The Quadro RTX $8000$ GPU used for this research was donated by the NVIDIA~Corporation.

\bibliography{bibtex-short}

@article{albumentations,
  title = {Albumentations: Fast and Flexible Image Augmentations},
  author = {Buslaev, Alexander and others},
  year = {2020},
  journal = {Information},
  volume = {11},
  number = {2},
  doi = {10.3390/info11020125},
  issn = {2078-2489},
  pages = {125},
}

@misc{caltech,
  title = {{Caltech Cars dataset}},
  author = {Markus Weber},
  year = {1999},
  howpublished = {\url{https://data.caltech.edu/records/20084}}
}

@misc{englishlp,
  title = {{EnglishLP database}},
  author = {Vlasta Srebri\'{c}},
  year = {2003},
  howpublished = {\url{https://www.zemris.fer.hr/projects/LicensePlates/english/baza_slika.zip}}
}

@misc{openalpr_eu,
  title = {{OpenALPR-EU dataset}},
  author = {{OpenALPR}},
  year = {2016},
  howpublished = {\url{https://github.com/openalpr/benchmarks/tree/master/endtoend/eu}}
}

@misc{rodosol,
  author = {{RodoSol}},
  title = {{\textit{Concessionária Rodovia do Sol S/A}}},
  howpublished = {\url{https://www.rodosol.com.br/blog/conheca-a-rodosol-2}},
  note = {Accessed: 2022-02-06},
  year = {2022},
}

@misc{ucsd,
  title = {{UCSD}/{C}alit2 Car License Plate, Make~and Model Database},
  author = {L. Dlagnekov and S. Belongie},
  year = {2005},
  howpublished = {\url{https://www.belongielab.org/car_data.html}}
}

@misc{openalprapi,
  title = {{OpenALPR Cloud API}},
  author = {{OpenALPR}},
  year = {2023},
  howpublished = {\url{http://www.openalpr.com/}}
}

@misc{sighthoundapi,
  title = {{Sighthound Cloud API}},
  author = {{Sighthound}},
  year = {2023},
  howpublished = {\url{https://www.sighthound.com/products/cloud}}
}

@inproceedings{mecocci2006generative,
  title = {Generative Models for License Plate Recognition by using a Limited Number of Training Samples},
  author = {Mecocci, Alessandro and Tommaso, Capasso},
  year = {2006},
  booktitle = {International Conference on Image Processing (ICIP)},
  volume = {},
  number = {},
  pages = {2769-2772},
  doi = {10.1109/ICIP.2006.313121}
}

@article{anagnostopoulos2008license,
  title = {License Plate Recognition From Still Images and Video Sequences: A Survey},
  author = {C. N. E. Anagnostopoulos and I. E. Anagnostopoulos and I. D. Psoroulas and V. Loumos and E. Kayafas},
  year = {2008},
  journal = {IEEE Transactions on Intelligent Transportation Systems},
  volume = {9},
  number = {3},
  pages = {377-391},
  doi = {10.1109/TITS.2008.922938},
  issn = {1524-9050}
}

@inproceedings{torralba2011unbiased,
  title = {Unbiased look at dataset bias},
  author = {Torralba, Antonio and Efros, Alexei A.},
  year = {2011},
  booktitle = {IEEE Conference on Computer Vision and Pattern Recognition (CVPR)},
  volume = {},
  number = {},
  pages = {1521-1528},
  doi = {10.1109/CVPR.2011.5995347}
}

@article{zhou2012principal,
  title = {Principal Visual Word Discovery for Automatic License Plate Detection},
  author = {W. Zhou and H. Li and Y. Lu and Q. Tian},
  year = {2012},
  month = {Sept},
  journal = {IEEE Transactions on Image Processing},
  volume = {21},
  number = {9},
  pages = {4269-4279},
  doi = {10.1109/TIP.2012.2199506},
  issn = {1057-7149}
}

@article{hsu2013application,
  title = {Application-Oriented License Plate Recognition},
  author = {G. S. Hsu and J. C. Chen and Y. Z. Chung},
  year = {2013},
  month = {Feb},
  journal = {IEEE Transactions on Vehicular Technology},
  volume = {62},
  number = {2},
  pages = {552-561},
  doi = {10.1109/TVT.2012.2226218},
  issn = {0018-9545},
}

@article{goncalves2016benchmark,
  title = {Benchmark for license plate character segmentation},
  author = {G. R. {Gon{\c{c}}alves} and da Silva, Sirlene Pio Gomes and Menotti, David and Schwartz, William Robson},
  year = {2016},
  journal = {Journal of Electronic Imaging},
  volume = {25},
  number = {5},
  pages = {053034},
  doi = {10.1117/1.JEI.25.5.053034},
  optpages = {053034-053034}
}

@INPROCEEDINGS{lee2016recursive,
  author={C. {Lee} and S. {Osindero}},
  booktitle={IEEE/CVF Conference on Computer Vision and Pattern Recognition~(CVPR)}, 
  title={Recursive Recurrent Nets with Attention Modeling for {OCR} in the Wild}, 
  year={2016},
  volume={},
  number={},
  pages={2231-2239},
  doi={10.1109/CVPR.2016.245}
}

@inproceedings{liu2016starnet,
  title={{STAR-Net}: A SpaTial Attention Residue Network for Scene Text Recognition},
  author={Wei {Liu} and Chaofeng Chen and Kwan-Yee K. Wong, Zhizhong Su and Junyu Han},
  year={2016},
  month={Sept},
  pages={1-13},
  booktitle={British Machine Vision Conference~(BMVC)},
  doi={},
  isbn={1-901725-59-6},
}

@INPROCEEDINGS{shi2016robust,
  author={B. {Shi} and X. {Wang} and P. {Lyu} and C. {Yao} and X. {Bai}},
  booktitle={IEEE/CVF Conference on Computer Vision and Pattern Recognition (CVPR)}, 
  title={Robust Scene Text Recognition with Automatic Rectification}, 
  year={2016},
  volume={},
  number={},
  pages={4168-4176},
  doi={10.1109/CVPR.2016.452}
}

@inproceedings{hsu2017robust,
  title = {Robust license plate detection in the wild},
  author = {G. S. Hsu and A. Ambikapathi and S. L. Chung and C. P. Su},
  year = {2017},
  month = {Aug},
  booktitle = {IEEE International Conference on Advanced Video and Signal Based Surveillance (AVSS)},
  volume = {},
  number = {},
  pages = {1-6},
  doi = {10.1109/AVSS.2017.8078493},
  issn = {}
}

@inproceedings{isola2017image,
  title = {Image-to-Image Translation with Conditional Adversarial Networks},
  author = {Phillip {Isola} and Zhu, Jun-Yan and Zhou, Tinghui and Efros, Alexei A.},
  year = {2017},
  booktitle = {IEEE Conference on Computer Vision and Pattern Recognition (CVPR)},
  volume = {},
  number = {},
  pages = {5967-5976},
  doi = {10.1109/CVPR.2017.632}
}

@article{panahi2017accurate,
  title = {Accurate Detection and Recognition of Dirty Vehicle Plate Numbers for High-Speed Applications},
  author = {R. Panahi and I. Gholampour},
  year = {2017},
  month = {April},
  journal = {IEEE Transactions on Intelligent Transportation Systems},
  volume = {18},
  number = {4},
  pages = {767-779},
  doi = {10.1109/TITS.2016.2586520},
  issn = {1524-9050}
}

@article{shi2017endtoend,
  title = {An End-to-End Trainable Neural Network for Image-Based Sequence Recognition and Its Application to Scene Text Recognition},
  author = {B. Shi and X. Bai and C. Yao},
  year = {2017},
  month = {Nov},
  journal = {IEEE Transactions on Pattern Analysis and Machine Intelligence},
  volume = {39},
  number = {11},
  pages = {2298-2304},
  doi = {10.1109/TPAMI.2016.2646371},
  issn = {0162-8828}
}

@inproceedings{spanhel2017holistic,
  title = {Holistic recognition of low quality license plates by {CNN} using track annotated data},
  author={J. {\v{S}pa\v{n}hel} and J. {Sochor} and R. {Jur\'{a}nek} and A. {Herout} and L. {Mar\v{s}\'{\i}k} and P. {Zem\v{c}\'{\i}k}},
  year = {2017},
  month = {Aug},
  booktitle = {IEEE International Conference on Advanced Video and Signal Based Surveillance},
  volume = {},
  number = {},
  pages = {1-6},
  doi = {10.1109/AVSS.2017.8078501},
  issn = {}
}

@article{yuan2017robust,
  title = {A Robust and Efficient Approach to License Plate Detection},
  author = {Y. {Yuan} and W. Zou and Y. Zhao and X. Wang and X. Hu and N. Komodakis},
  year = {2017},
  month = {March},
  journal = {IEEE Transactions on Image Processing},
  volume = {26},
  number = {3},
  pages = {1102-1114},
  doi = {10.1109/TIP.2016.2631901},
  issn = {1057-7149}
}

@inproceedings{wang2017gated,
  title = {Gated Recurrent Convolution Neural Network for {OCR}},
  author = {Jianfeng {Wang} and Hu, Xiaolin},
  year = {2017},
  booktitle = {International Conf. on Neural Information Processing Systems (NeurIPS)},
  pages = {334–343},
  isbn = {9781510860964},
  doi = {10.5555/3294771.3294803},
}

@inproceedings{zhu2017unpaired,
  title = {Unpaired Image-to-Image Translation Using Cycle-Consistent Adversarial Networks},
  author = {Jun-Yan {Zhu} and Park, Taesung and Isola, Phillip and Efros, Alexei A.},
  year = {2017},
  booktitle = {IEEE International Conference on Computer Vision (ICCV)},
  volume = {},
  number = {},
  pages = {2242-2251},
  doi = {10.1109/ICCV.2017.244}
}

@inproceedings{borisyuk2018rosetta,
  author = {Borisyuk, Fedor and Gordo, Albert and Sivakumar, Viswanath},
  title = {Rosetta: Large Scale System for Text Detection and Recognition in Images},
  year = {2018},
  isbn = {9781450355520},
  booktitle = {ACM SIGKDD International Conference on Knowledge Discovery \& Data Mining},
  pages = {71–79},
  doi = {10.1145/3219819.3219861},
}

@inproceedings{goncalves2018realtime,
  title = {Real-Time Automatic License Plate Recognition through Deep Multi-Task Networks},
  author = {G. R. {Gon{\c{c}}alves} and M. A. {Diniz} and R. {Laroca} and D. {Menotti} and W. R. {Schwartz}},
  year = {2018},
  month = {Oct},
  booktitle = {Conference on Graphics, Patterns and Images (SIBGRAPI)},
  volume = {},
  number = {},
  pages = {110-117},
  doi = {10.1109/SIBGRAPI.2018.00021},
  issn = {2377-5416}
}

@inproceedings{laroca2018robust,
  title = {A Robust Real-Time Automatic License Plate Recognition Based on the {YOLO} Detector},
  author = {R. {Laroca} and others},
  year = {2018},
  month = {July},
  booktitle = {International Joint Conference on Neural Networks (IJCNN)},
  volume = {},
  number = {},
  pages = {1-10},
  doi = {10.1109/IJCNN.2018.8489629},
  issn = {2161-4407}
}

@article{li2018reading,
  title = {Reading car license plates using deep neural networks},
  author = {Hui {Li} and Peng Wang and Mingyu You and Chunhua Shen},
  year = {2018},
  journal = {Image and Vision Computing},
  volume = {72},
  pages = {14-23},
  doi = {10.1016/j.imavis.2018.02.002},
  issn = {0262-8856}
}

@inproceedings{meng2018robust,
  title = {A robust and efficient method for license plate recognition},
  author = {A. {Meng} and W. {Yang} and Z. {Xu} and H. {Huang} and L. {Huang} and C. {Ying}},
  year = {2018},
  month = {Aug},
  booktitle = {International Conference on Pattern Recognition~(ICPR)},
  volume = {},
  number = {},
  pages = {1713-1718},
  doi = {10.1109/ICPR.2018.8546291},
  issn = {1051-4651}
}

@inproceedings{silva2018license,
  title = {License Plate Detection and Recognition in Unconstrained Scenarios},
  author = {S. M. {Silva} and C. R. {Jung}},
  year = {2018},
  month = {Sept},
  booktitle = {European Conf. on Computer Vision (ECCV)},
  pages = {593-609},
  doi = {10.1007/978-3-030-01258-8_36}
}

@article{xie2018new,
  title = {A New {CNN}-Based Method for Multi-Directional Car License Plate Detection},
  author = {L. Xie and T. Ahmad and L. Jin and Y. Liu and S. Zhang},
  year = {2018},
  month = {Feb},
  journal = {IEEE Transactions on Intelligent Transportation Systems},
  volume = {19},
  number = {2},
  pages = {507-517},
  doi = {10.1109/TITS.2017.2784093},
  issn = {1524-9050}
}

@inproceedings{xu2018towards,
  title = {Towards End-to-End License Plate Detection and Recognition: A Large Dataset and Baseline},
  author = {Xu, Zhenbo and Yang, Wei and Meng, Ajin and Lu, Nanxue and Huang, Huan and Ying, Changchun and Huang, Liusheng},
  year = {2018},
  booktitle = {European Conference on Computer Vision (ECCV)},
  pages = {261-277},
  doi = {10.1007/978-3-030-01261-8_16},
  isbn = {978-3-030-01261-8}
}

@article{zhang2018vehicle,
  title = {Vehicle license plate detection and recognition using deep neural networks and generative adversarial networks},
  author = {Xiaoci {Zhang} and Naijie Gu and Hong Ye and Chuanwen Lin},
  year = {2018},
  journal = {Journal of Electronic Imaging},
  volume = {27},
  number = {4},
  pages = {043056},
  doi = {10.1117/1.JEI.27.4.043056}
}

@inproceedings{zhuang2018towards,
  title = {Towards Human-Level License Plate Recognition},
  author = {Jiafan {Zhuang} and others},
  year = {2018},
  booktitle = {European Conference on Computer Vision (ECCV)},
  pages = {314-329},
  doi = {10.1007/978-3-030-01219-9_19},
  isbn = {978-3-030-01219-9}
}

@inproceedings{baek2019what,
  title = {What Is Wrong With Scene Text Recognition Model Comparisons? Dataset and Model Analysis},
  author = {Jeonghun {Baek} and Kim, Geewook and Lee, Junyeop and Park, Sungrae and Han, Dongyoon and Yun, Sangdoo and Oh, Seong Joon and Lee, Hwalsuk},
  year = {2019},
  booktitle = {IEEE/CVF International Conference on Computer Vision (ICCV)},
  volume = {},
  number = {},
  pages = {4714-4722},
  doi = {10.1109/ICCV.2019.00481}
}

@article{bjorklund2019robust,
  title = {Robust license plate recognition using neural networks trained on synthetic images},
  author = {Tomas Bj\"{o}rklund and Attilio Fiandrotti and Mauro Annarumma and Gianluca Francini and Enrico Magli},
  year = {2019},
  journal = {Pattern Recognition},
  volume = {93},
  pages = {134-146},
  doi = {10.1016/j.patcog.2019.04.007},
  issn = {0031-3203}
}

@inproceedings{goncalves2019multitask,
  title = {Multi-Task Learning for Low-Resolution License Plate Recognition},
  author = {G. R. {Gon{\c{c}}alves} and others},
  year = {2019},
  month = {Oct},
  booktitle = {Iberoamerican Congress on Pattern Recognition},
  volume = {},
  number = {},
  pages = {251-261},
  doi = {10.1007/978-3-030-33904-3_23}
}

@article{laroca2019convolutional,
  author = {R. {Laroca} and V. {Barroso} and M. A. {Diniz} and G. R. {Gon{\c{c}}alves} and W. R. {Schwartz} and D. {Menotti}},
  title = {Convolutional Neural Networks for Automatic Meter Reading},
  journal = {Journal of Electronic Imaging},
  volume = {28},
  number = {1},
  pages = {013023},
  year = {2019},
  ISSN = {1017-9909},
  doi = {10.1117/1.JEI.28.1.013023},
}

@article{li2019toward,
  title = {Toward End-to-End Car License Plate Detection and Recognition With Deep Neural Networks},
  author = {Hui {Li} and Wang, Peng and Shen, Chunhua},
  year = {2019},
  journal = {IEEE Transactions on Intelligent Transportation Systems},
  volume = {20},
  number = {3},
  pages = {1126-1136},
  doi = {10.1109/TITS.2018.2847291}
}

@article{wu2019pixtextgan,
  title = {{PixTextGAN}: structure aware text image synthesis for license plate recognition},
  author = {Wu, Shilian and Zhai, Wei and Cao, Yang},
  year = {2019},
  journal = {IET Image Processing},
  volume = {13},
  number = {14},
  pages = {2744--2752},
  doi = {10.1049/iet-ipr.2018.6588}
}

@article{xiang2019lightweight,
  title = {Lightweight fully convolutional network for license plate detection},
  author = {Han Xiang and Yong Zhao and Yule Yuan and Guiying Zhang and Xuefeng Hu},
  year = {2019},
  journal = {Optik},
  volume = {178},
  pages = {1185-1194},
  doi = {10.1016/j.ijleo.2018.10.098},
  issn = {0030-4026}
}

@article{castrozunti2020license,
  title = {License plate segmentation and recognition system using deep learning and {OpenVINO}},
  author = {Riel D. {Castro-Zunti} and Y\'{e}pez, Juan and Ko, Seok-Bum},
  year = {2020},
  journal = {IET Intelligent Transport Systems},
  volume = {14},
  number = {2},
  pages = {119--126},
  doi = {10.1049/iet-its.2019.0481}
}

@inproceedings{chan2020european,
  title = {{European Union} Dataset and Annotation Tool for Real Time Automatic License Plate Detection and Blurring},
  author = {Lap Yan {Chan} and others},
  year = {2020},
  booktitle = {IEEE International Conference on Intelligent Transportation Systems (ITSC)},
  volume = {},
  number = {},
  pages = {1--6},
  doi = {10.1109/ITSC45102.2020.9294240}
}

@article{han2020license,
  title = {License Plate Image Generation using Generative Adversarial Networks for End-To-End License Plate Character Recognition from a Small Set of Real Images},
  author = {Han, Byung-Gil and Lee, Jong Taek and Lim, Kil-Taek and Choi, Doo-Hyun},
  year = {2020},
  journal = {Applied Sciences},
  volume = {10},
  number = {8},
  pages = {2780},
  doi = {10.3390/app10082780},
  issn = {2076-3417}
}

@article{henry2020multinational,
  title = {Multinational License Plate Recognition Using Generalized Character Sequence Detection},
  author = {C. {Henry} and S. Y. {Ahn} and S. {Lee}},
  year = {2020},
  journal = {IEEE Access},
  volume = {8},
  number = {},
  pages = {35185-35199},
  doi = {10.1109/ACCESS.2020.2974973}
}

@article{hu2020mobilenet,
  title = {{MobileNet-SSD MicroScope} using adaptive error correction algorithm: real-time detection of license plates on mobile devices},
  author = {X. {Hu} and H. {Li} and X. {Li} and C. {Wang}},
  year = {2020},
  journal = {IET Intelligent Transport Systems},
  volume = {14},
  number = {2},
  pages = {110-118},
  doi = {10.1049/iet-its.2019.0380}
}

@article{silva2020realtime,
  title = {Real-Time License Plate Detection and Recognition Using Deep Convolutional Neural Networks},
  author = {S. M. {Silva} and C. R. {Jung}},
  year = {2020},
  journal = {Journal of Visual Communication and Image Representation},
  pages = {102773},
  doi = {10.1016/j.jvcir.2020.102773},
  issn = {1047-3203}
}

@article{weihong2020research,
  title = {Research on License Plate Recognition Algorithms Based on Deep Learning in Complex Environment},
  author = {W. {Weihong} and T. {Jiaoyang}},
  year = {2020},
  journal = {IEEE Access},
  volume = {8},
  number = {},
  pages = {91661-91675},
  doi = {10.1109/ACCESS.2020.2994287}
}

@article{yoo2021deep,
  title = {Deep corner prediction to rectify tilted license plate images},
  author = {Hojin {Yoo} and Kyungkoo Jun},
  year = {2021},
  journal = {Multimedia Systems},
  volume = {27},
  number = {4},
  pages = {779-786},
  doi = {10.1007/s00530-020-00655-8}
}

@article{zhang2021robust_attentional,
  title = {A Robust Attentional Framework for License Plate Recognition in the Wild},
  author = {Linjiang {Zhang} and Wang, Peng and Li, Hui and Li, Zhen and Shen, Chunhua and Zhang, Yanning},
  year = {2021},
  journal = {IEEE Transactions on Intelligent Transportation Systems},
  volume={22},
  number={11},
  pages={6967-6976},
  doi = {10.1109/TITS.2020.3000072}
}

@article{zou2020robust,
  title = {A Robust License Plate Recognition Model Based on {Bi-LSTM}},
  author = {Zou, Yongjie and others},
  year = {2020},
  journal = {IEEE Access},
  volume = {8},
  number = {},
  pages = {211630-211641},
  doi = {10.1109/ACCESS.2020.3040238}
}

@article{liu2024improving,
  title = {Improving Multi-Type License Plate Recognition via Learning Globally and Contrastively},
  author = {Qi {Liu} and Liu, Yan and Chen, Song-Lu and Zhang, Tian-Hao and Chen, Feng and Yin, Xu-Cheng},
  year = {2024},
  journal = {IEEE Transactions on Intelligent Transportation Systems},
  volume = {25},
  number = {9},
  pages = {11092-11102},
  doi = {10.1109/TITS.2024.3365537}
}

@inproceedings{atienza2021vitstr,
  title = {Vision Transformer for Fast and Efficient Scene Text Recognition},
  author = {Rowel {Atienza}},
  year = {2021},
  booktitle = {International Conf. on Document Analysis and Recognition},
  pages = {319-334},
  doi = {10.1007/978-3-030-86549-8_21},
  isbn = {978-3-030-86549-8}
}

@inproceedings{baek2021what,
  title = {What If We Only Use Real Datasets for Scene Text Recognition? Toward Scene Text Recognition With Fewer Labels},
  author = {Jeonghun {Baek} and Matsui, Yusuke and Aizawa, Kiyoharu},
  year = {2021},
  booktitle = {IEEE/CVF Conference on Computer Vision and Pattern Recognition (CVPR)},
  volume = {},
  number = {},
  pages = {3112-3121},
  doi = {10.1109/CVPR46437.2021.00313}
}

@article{beratoglu2021vehicle,
  title = {Vehicle License Plate Detector in Compressed Domain},
  author = {Berato\u{g}lu, Muhammet Sebul and T\"{o}rey\.{i}n, Beh\c{c}et U\u{g}ur},
  year = {2021},
  journal = {IEEE Access},
  volume = {9},
  number = {},
  pages = {95087-95096},
  doi = {10.1109/ACCESS.2021.3092938}
}

@article{kong2021federated,
  title = {A Federated Learning-Based License Plate Recognition Scheme for {5G}-Enabled Internet of Vehicles},
  author = {Xiangjie {Kong} and Wang, Kailai and Hou, Mingliang and Hao, Xinyu and Shen, Guojiang and Chen, Xin and Xia, Feng},
  year = {2021},
  journal = {IEEE Transactions on Industrial Informatics},
  volume = {17},
  number = {12},
  pages = {8523-8530},
  doi = {10.1109/TII.2021.3067324}
}

@article{laroca2021efficient,
  title = {An Efficient and Layout-Independent Automatic License Plate Recognition System Based on the {YOLO} Detector},
  author = {R. {Laroca} and L. A. {Zanlorensi} and G. R. {Gon{\c{c}}alves} and E. {Todt} and W. R. {Schwartz} and D. {Menotti}},
  year = {2021},
  journal = {IET Intelligent Transport Systems},
  volume = {15},
  number = {4},
  pages = {483-503},
  doi = {10.1049/itr2.12030},
  issn = {1751-956X}
}

@article{laroca2021towards,
  title = {Towards Image-Based Automatic Meter Reading in Unconstrained Scenarios: A Robust and Efficient Approach},
  author = {R. {Laroca} and A. B. {Araujo} and L. A. {Zanlorensi} and E. C. {De Almeida} and D. {Menotti}},
  year = {2021},
  journal = {IEEE Access},
  volume = {9},
  number = {},
  pages = {67569-67584},
  doi = {10.1109/ACCESS.2021.3077415},
  issn = {2169-3536}
}

@article{qiao2021mango,
  title = {{MANGO}: A Mask Attention Guided One-Stage Scene Text Spotter},
  author = {Liang {Qiao} and Chen, Ying and Cheng, Zhanzhan and Xu, Yunlu and Niu, Yi and Pu, Shiliang and Wu, Fei},
  year = {2021},
  month = {May},
  journal = {AAAI Conference on Artificial Intelligence},
  volume = {35},
  number = {3},
  pages = {2467-2476},
 }

@inproceedings{liu2021fast,
  title = {Fast Recognition for Multidirectional and Multi-type License Plates with {2D} Spatial Attention},
  author = {Qi {Liu} and Chen, Song-Lu and Li, Zhen-Jia and Yang, Chun and Chen, Feng and Yin, Xu-Cheng},
  year = {2021},
  booktitle = {International Conference on Document Analysis and Recognition (ICDAR)},
  pages = {125--139},
  doi = {10.1007/978-3-030-86337-1_9},
  isbn = {978-3-030-86337-1}
}

@article{lubna2021automatic,
  title = {Automatic Number Plate {Recognition}: A Detailed Survey of Relevant Algorithms},
  author = {Lubna and Mufti, Naveed and Shah, Syed Afaq Ali},
  year = {2021},
  journal = {Sensors},
  volume = {21},
  number = {9},
  pages = {3028},
  doi = {10.3390/s21093028},
  issn = {1424-8220}
}

@inproceedings{wang2021scaledyolov4,
  title = {{Scaled-YOLOv4}: Scaling Cross Stage Partial Network},
  author = {Chien-Yao {Wang} and Bochkovskiy, Alexey and Liao, Hong-Yuan Mark},
  year = {2021},
  month = {June},
  booktitle = {IEEE/CVF Conference on Computer Vision and Pattern Recognition},
  pages = {13029-13038},
  doi={10.1109/CVPR46437.2021.01283}
}

@inproceedings{zhang2021efficient,
  title = {Efficient License Plate Recognition via Holistic Position Attention},
  author = {Y. {Zhang} and others},
  year = {2021},
  month = {May},
  booktitle = {AAAI Conference on Artificial Intelligence},
  volume = {},
  number = {},
  pages = {3438-3446},
  doi = {10.1609/aaai.v35i4.16457},
}

@article{zhang2021vlpdr,
  title = {{V-LPDR}: Towards a unified framework for license plate detection, tracking, and recognition in real-world traffic videos},
  author = {C. {Zhang} and Q. {Wang} and X. {Li}},
  year = {2021},
  journal = {Neurocomputing},
  volume = {449},
  pages = {189-206},
  doi = {10.1016/j.neucom.2021.03.103},
  issn = {0925-2312}
}

@article{albatat2022end,
  title = {An End-to-End Automated License Plate Recognition System Using {YOLO} Based Vehicle and License Plate Detection with Vehicle Classification},
  author = {Reda {Al-batat} and Angelopoulou, Anastassia and Premkumar, Smera and Hemanth, Jude and Kapetanios, Epameinondas},
  year = {2022},
  journal = {Sensors},
  volume = {22},
  number = {23},
  pages = {9477},
  doi = {10.3390/s22239477},
  issn = {1424-8220}
}

@article{fan2022improving,
  title = {Improving Robustness of License Plates Automatic Recognition in Natural Scenes},
  author = {Xudong {Fan} and Zhao, Wei},
  year = {2022},
  journal = {IEEE Transactions on Intelligent Transportation Systems},
  volume = {23},
  number = {10},
  pages = {18845-18854},
  doi = {10.1109/TITS.2022.3151475}
}

@inproceedings{laroca2022cross,
  title = {On the Cross-dataset Generalization in License Plate Recognition},
  author = {R. {Laroca} and E. V. {Cardoso} and D. R. {Lucio} and V. {Estevam} and D. {Menotti}},
  year = {2022},
  month = {Feb},
  booktitle = {International Conference on Computer Vision Theory and Applications (VISAPP)},
  volume = {},
  number = {},
  pages = {166-178},
  doi = {10.5220/0010846800003124},
  isbn = {978-989-758-555-5},
  issn = {2184-4321}
}

@inproceedings{laroca2022first,
  title = {A First Look at Dataset Bias in License Plate Recognition},
  author = {R. {Laroca} and M. {Santos} and V. {Estevam} and E. {Luz} and D. {Menotti}},
  year = {2022},
  month = {Oct},
  booktitle = {Conference on Graphics, Patterns and Images (SIBGRAPI)},
  volume = {},
  number = {},
  pages = {234-239},
  doi = {10.1109/SIBGRAPI55357.2022.9991768},
  issn = {1530-1834}
}

@article{liang2022egsanet,
  title = {{EGSANet}: edge-guided sparse attention network for improving license plate detection in the wild},
  author = {Jing {Liang} and Chen, Guancheng and Wang, Yan and Qin, Huabiao},
  year = {2022},
  journal = {Applied Intelligence},
  volume = {52},
  number = {4},
  pages = {4458-4472},
  doi = {10.1007/s10489-021-02628-4},
  issn = {1573-7497}
}

@inproceedings{maier2022reliability,
  title = {Reliability Scoring for the Recognition of Degraded License Plates},
  author = {Anatol {Maier} and Moussa, Denise and Spruck, Andreas and Seiler, J\"{u}rgen and Riess, Christian},
  year = {2022},
  booktitle = {IEEE International Conference on Advanced Video and Signal Based Surveillance (AVSS)},
  volume = {},
  number = {},
  pages = {1-8},
  doi = {10.1109/AVSS56176.2022.9959390}
}

@article{shashirangana2022license,
  title = {License plate recognition using neural architecture search for edge devices},
  author = {Jithmi {Shashirangana} and Padmasiri, Heshan and Meedeniya, Dulani and Perera, Charith and Nayak, Soumya R. and Nayak, Janmenjoy and Vimal, Shanmuganthan and Kadry, Seifidine},
  year = {2022},
  journal = {International Journal of Intelligent Systems},
  volume = {37},
  number = {12},
  pages = {10211-10248},
  doi = {10.1002/int.22471}
}

@article{silva2022flexible,
  title = {A Flexible Approach for Automatic License Plate Recognition in Unconstrained Scenarios},
  author = {S. M. {Silva} and C. R. {Jung}},
  year = {2022},
  journal = {IEEE Transactions on Intelligent Transportation Systems},
  volume = {23},
  number = {6},
  pages = {5693-5703},
  doi = {10.1109/TITS.2021.3055946}
}

@inproceedings{trinh2023pp4av,
  title = {{PP4AV}: A Benchmarking Dataset for Privacy-Preserving Autonomous Driving},
  author = {Linh {Trinh} and Pham, Phuong and Trinh, Hoang and Bach, Nguyen and Nguyen, Dung and Nguyen, Giang and Nguyen, Huy},
  year = {2023},
  booktitle = {IEEE/CVF Winter Conference on Applications of Computer Vision (WACV)},
  volume = {},
  number = {},
  pages = {1206-1215},
  doi = {10.1109/WACV56688.2023.00126}
}

@inproceedings{wang2022efficient,
  title = {Efficient License Plate Recognition via Parallel Position-Aware Attention},
  author = {Tianxiang {Wang} and others},
  year = {2022},
  booktitle = {Pattern Recognition and Computer Vision},
  pages = {346-360},
  doi = {10.1007/978-3-031-18913-5_27},
  isbn = {978-3-031-18913-5}
}

@article{wang2022rethinking,
  title = {Rethinking and Designing a High-Performing Automatic License Plate Recognition Approach},
  author = {Yi {Wang} and Bian, Zhen-Peng and Zhou, Yunhao and Chau, Lap-Pui},
  year = {2022},
  journal = {IEEE Transactions on Intelligent Transportation Systems},
  volume = {23},
  number = {7},
  pages = {8868-8880},
  doi = {10.1109/TITS.2021.3087158}
}

@article{xu2022eilpr,
  title = {{EILPR}: Toward End-to-End Irregular License Plate Recognition Based on Automatic Perspective Alignment},
  author = {Hui {Xu} and Zhou, Xiang-Dong and Li, Zhenghao and Liu, Liangchen and Li, Chaojie and Shi, Yu},
  year = {2022},
  journal = {IEEE Transactions on Intelligent Transportation Systems},
  volume = {23},
  number = {3},
  pages = {2586-2595},
  doi = {10.1109/TITS.2021.3130898}
}

@article{zou2022license,
  title = {License plate detection and recognition based on {YOLOv3} and {ILPRNET}},
  author = {Yongjie {Zou} and Zhang, Yongjun and Yan, Jun and Jiang, Xiaoxu and Huang, Tengjie and Fan, Haisheng and Cui, Zhongwei},
  year = {2022},
  month = {},
  day = {},
  journal = {Signal, Image and Video Processing},
  volume = {16},
  number = {2},
  pages = {473-480},
  doi = {10.1007/s11760-021-01981-8},
  issn = {1863-1711}
}

@inproceedings{chen2023endtoend,
  title = {End-to-End Multi-line License Plate Recognition with Cascaded Perception},
  author = {Song-Lu {Chen} and Liu, Qi and Chen, Feng and Yin, Xu-Cheng},
  year = {2023},
  booktitle = {International Conference on Document Analysis and Recognition (ICDAR)},
  pages = {274-289},
  doi = {10.1007/978-3-031-41734-4_17},
  isbn = {978-3-031-41734-4}
}

@article{gao2023group,
  title = {{GroupPlate}: Toward Multi-Category License Plate Recognition},
  author = {Yilin {Gao} and others},
  year = {2023},
  journal = {IEEE Transactions on Intelligent Transportation Systems},
  volume = {24},
  number = {5},
  pages = {5586-5599},
  doi = {10.1109/TITS.2023.3244827}
}

@article{jia2023efficient,
  title = {An Efficient License Plate Detection Approach With Deep Convolutional Neural Networks in Unconstrained Scenarios},
  author = {Wei {Jia} and Xie, Mingshan},
  year = {2023},
  journal = {IEEE Access},
  volume = {11},
  number = {},
  pages = {85626-85639},
  doi = {10.1109/ACCESS.2023.3301122}
}

@article{jiang2023efficient,
  title = {An Efficient and Unified Recognition Method for Multiple License Plates in Unconstrained Scenarios},
  author = {Yu {Jiang} and Jiang, Feng and Luo, Huiyin and Lin, Hongyu and Yao, Jian and Liu, Jiaxin and Ren, Jia},
  year = {2023},
  journal = {IEEE Transactions on Intelligent Transportation Systems},
  volume = {24},
  number = {5},
  pages = {5376-5389},
  doi = {10.1109/TITS.2023.3237743}
}

@article{ke2023ultra,
  title = {An Ultra-Fast Automatic License Plate Recognition Approach for Unconstrained Scenarios},
  author = {Xiao {Ke} and Zeng, Ganxiong and Guo, Wenzhong},
  year = {2023},
  journal = {IEEE Transactions on Intelligent Transportation Systems},
  volume = {24},
  number = {5},
  pages = {5172-5185},
  doi = {10.1109/TITS.2023.3237581}
}

@inproceedings{laroca2023do,
  title = {Do We Train on Test Data? {T}he Impact of Near-Duplicates on License Plate Recognition},
  author = {R. {Laroca} and V. {Estevam} and A. S. {Britto Jr.} and R. {Minetto} and D. {Menotti}},
  year = {2023},
  month = {Jun},
  booktitle = {International Joint Conference on Neural Networks (IJCNN)},
  volume = {},
  number = {},
  pages = {1-8},
  doi = {10.1109/IJCNN54540.2023.10191584}
}

@inproceedings{laroca2023leveraging,
  title = {Leveraging Model Fusion for Improved License Plate Recognition},
  author = {R. {Laroca} and L. A. {Zanlorensi} and V. {Estevam} and R. {Minetto} and D. {Menotti}},
  year = {2023},
  month = {Nov},
  booktitle = {Iberoamerican Congress on Pattern Recognition (CIARP)},
  volume = {},
  number = {},
  pages = {60-75},
  doi = {10.1007/978-3-031-49249-5_5},
  isbn = {978-3-031-49249-5}
}

@article{nascimento2023super,
  title = {Super-Resolution of License Plate Images Using Attention Modules and Sub-Pixel Convolution Layers},
  author = {V. {Nascimento} and others},
  year = {2023},
  journal = {Computers \& Graphics},
  volume = {113},
  number = {},
  pages = {69-76},
  doi = {10.1016/j.cag.2023.05.005},
  issn = {0097-8493}
}

@article{schirrmacher2023benchmarking,
  title = {Benchmarking Probabilistic Deep Learning Methods for License Plate Recognition},
  author = {Franziska {Schirrmacher} and Lorch, Benedikt and Maier, Anatol and Riess, Christian},
  year = {2023},
  journal = {IEEE Transactions on Intelligent Transportation Systems},
  volume = {24},
  number = {9},
  pages = {9203-9216},
  doi = {10.1109/TITS.2023.3278533}
}

@article{shvai2023multiple,
  title = {Multiple auxiliary classifiers {GAN} for controllable image generation: Application to license plate recognition},
  author = {Nadiya {Shvai} and Hasnat, Abul and Nakib, Amir},
  year = {2023},
  journal = {IET Intelligent Transport Systems},
  volume = {17},
  number = {1},
  pages = {243-254},
  doi = {10.1049/itr2.12251}
}

@article{zhou2023fafenet,
  title = {{FAFEnet}: A fast and accurate model for automatic license plate detection and recognition},
  author = {Xin {Zhou} and others},
  year = {2023},
  journal = {IET Image Processing},
  volume = {17},
  number = {3},
  pages = {807-818},
  doi = {10.1049/ipr2.12674}
}

@article{ding2024endtoend,
  title        = {An End-to-End Contrastive License Plate Detector},
  author       = {Haoxuan {Ding} and Gao, Junyu and Yuan, Yuan and Wang, Qi},
  year         = {2024},
  journal      = {IEEE Transactions on Intelligent Transportation Systems},
  volume       = {25},
  number       = {1},
  pages        = {503-516},
  doi          = {10.1109/TITS.2023.3304816}
}

@article{rao2024license,
  title = {License plate recognition system in unconstrained scenes via a new image correction scheme and improved {CRNN}},
  author = {Zhan {Rao} and Dezhi Yang and Ning Chen and Jian Liu},
  year = {2024},
  journal = {Expert Systems with Applications},
  volume = {243},
  pages = {122878},
  doi = {10.1016/j.eswa.2023.122878},
  issn = {0957-4174}
}

@inproceedings{trinh2022layout,
  title = {Layout-invariant license plate detection and recognition},
  author = {Thi-Anh-Loan {Trinh} and Pham, The-Anh and Hoang, Van-Dung},
  year = {2022},
  booktitle = {International Conference on Multimedia Analysis and Pattern Recognition},
  volume = {},
  number = {},
  pages = {1-6},
  doi = {10.1109/MAPR56351.2022.9924802}
}

@inproceedings{liu2024irregular,
  title = {Irregular License Plate Recognition via Global Information Integration},
  author = {Yuan-Yuan {Liu} and others},
  year = {2024},
  booktitle = {International Conf. on Multimedia Modeling},
  pages = {325-339},
  doi = {10.1007/978-3-031-53308-2_24},
  isbn = {978-3-031-53308-2}
}

\end{document}